\documentclass{article}

\PassOptionsToPackage{numbers}{natbib}

\usepackage[main, final]{neurips_2026}

\usepackage[utf8]{inputenc}
\usepackage[T1]{fontenc}
\usepackage{hyperref}
\usepackage{url}
\usepackage{booktabs}
\usepackage{amsfonts}
\usepackage{nicefrac}
\usepackage{microtype}
\usepackage{xcolor}

\usepackage{hyperref}
\usepackage{url}
\usepackage{multicol}
\usepackage{multirow}
\usepackage{subcaption}
\usepackage{amsmath}
\usepackage{graphicx}
\usepackage{subcaption}
\usepackage{caption}
\usepackage{wrapfig}
\usepackage{booktabs}
\usepackage{booktabs}
\usepackage{longtable}
\usepackage{geometry}
\title{AbFlowNet: Optimizing Antibody-Antigen Binding Energy via Diffusion-GFlowNet Fusion}

\author{Abrar Rahman Abir$^{1}$\thanks{Equal contribution; ordered alphabetically.}, Haz Sameen Shahgir$^{2}$\footnotemark[1]\\
\textbf{Md. Rownok Zahan Ratul$^3$}, \textbf{Md Toki Tahmid$^4$}, \textbf{Greg Ver Steeg$^2$}, \textbf{Yue Dong$^2$}\\ [5pt]
University of California, Berkeley $^1$, University of California, Riverside $^2$,\\ University of Southern California$^3$, Princeton University$^4$\\ [5pt]
\texttt{abir@berkeley.edu}, \texttt{hshah057@ucr.edu}
}

\begin{document}

\maketitle

\begin{abstract}
Complementarity Determining Regions (CDRs) are critical segments of an antibody that facilitate binding to specific antigens. Current computational methods for CDR design rely on reconstruction losses and do not jointly optimize binding energy, a crucial metric of antibody efficacy. Rather, binding energy optimization is performed via computationally expensive Online Reinforcement Learning (RL) pipelines that rely heavily on unreliable binding-energy estimators. In this paper, we propose AbFlowNet, a novel generative framework that integrates GFlowNet with Diffusion models. By framing each diffusion step as a state in the GFlowNet framework, AbFlowNet jointly optimizes standard diffusion losses and binding energy by directly incorporating energy signals into the training process, thereby unifying diffusion and reward optimization in a single procedure. Experimental results show that AbFlowNet outperforms the base diffusion model by $3.06\%$ in amino acid recovery, $20.40\%$ in geometric reconstruction (RMSD), and $3.60\%$ in binding energy improvement ratio. ABFlowNet also decreases Top-1 total energy and binding energy errors by $24.8\%$ and $38.1\%$, which is competitive with AbDPO's expensive online RL optimization without pseudo-labeling the test dataset or requiring computationally expensive synthetic CDR sampling. \footnote{The code and model weights are available at
\url{https://github.com/Patchwork53/abflownet}.}
\end{abstract}

\section{Introduction}

Antibodies are central to adaptive immunity, and their complementarity-determining regions (CDRs) largely determine antigen recognition and binding specificity. Because CDRs are highly variable and directly shape antigen binding \citep{cdr_basics}, they are a primary target for therapeutic antibody engineering, including CDR grafting and shuffling in antibody humanization \citep{cdr_grafting,cdr_shuffling,cdr_use1,cdr_use2}.

Designing \textit{de novo} CDRs for novel antigens \citep{tang2024surveygenerativeainovo} and neoantigens \citep{zhang2021neoantigen} remains challenging because the search space is combinatorial: a CDR of length $L$ admits $20^L$ sequences before structural variation. Classical Monte Carlo approaches directly optimize biophysical energy functions \citep{adolf2018rosettaantibodydesign,lapidoth2015abdesign,rabd}, but are computationally expensive and prone to local optima \citep{hern}. Recent deep learning approaches, including GNN-based \citep{abgnn,hern,dymean,mean} and diffusion-based models \citep{diffab,abdesign_abdock,abx}, help navigate this space by learning distributions of plausible CDR sequences and structures. However, diffusion models are typically trained with reconstruction-style objectives, which improve generative fidelity but do not directly optimize binding affinity.

\begin{figure*}[th]
    \centering
    \includegraphics[width=1\linewidth]{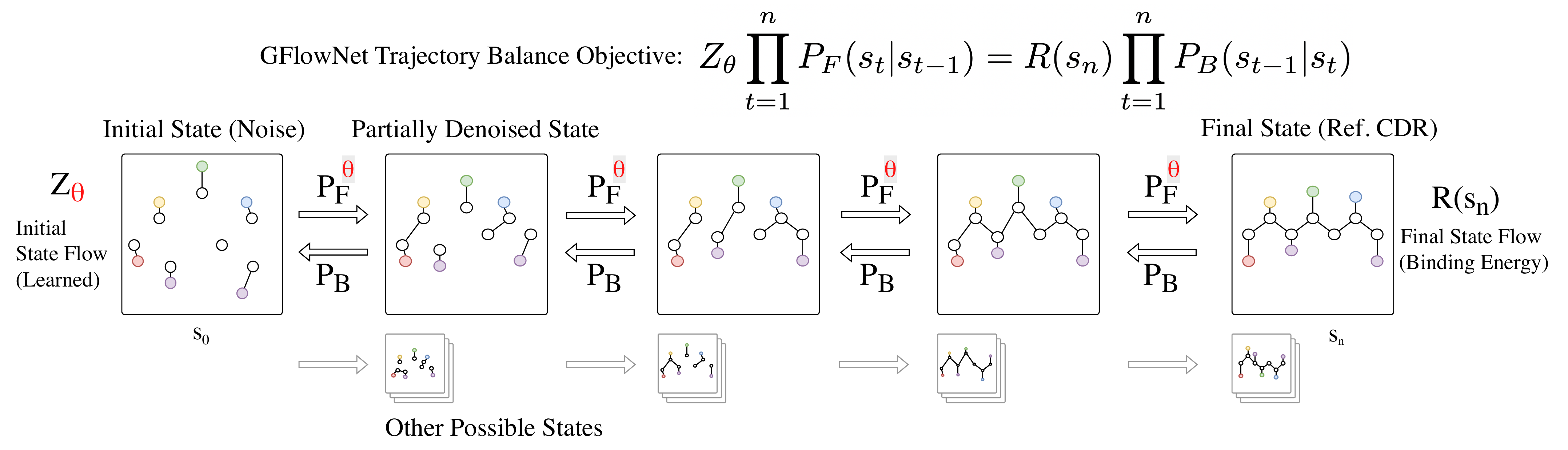}
    \caption{
     AbFlowNet reframes the diffusion process as a GFlowNet where each partially denoised CDR is a \textit{state} and the transition probabilities are \textit{flows through edges}. The initial state's flow is learned, and the final state's flow is the binding energy of the reference CDR. To train, we simply enforce forward and backward flow parity, in addition to the diffusion losses.
     }
    \vspace{-1.5em}
    \label{fig:openfig}
\end{figure*}

This creates a central challenge for affinity optimization in diffusion-based antibody generation: binding-affinity supervision is sparse and usually available only for the final generated structure. Recent work addresses this challenge by coupling diffusion models with online reinforcement learning, using energy estimators as rewards \citep{sutton1998reinforcement,abdpo,abnovo,alignab}. However, online RL is computationally expensive, depends on noisy energy estimates \citep{binding_energy_survey}, may introduce bias when test complexes are used during RL training \citep{abdpo,alignab,contamination}, and can degrade structural fidelity metrics such as Amino Acid Recovery and RMSD \citep{abdpo}. While recent antibody generation models increasingly use pretrained protein models and stronger architectures to improve structural plausibility and sequence realism, these advances mainly improve the generative backbone. Our focus is instead the optimization objective: efficiently propagating sparse terminal binding-affinity rewards through intermediate denoising steps during diffusion training.

Given the limitations of RL, GFlowNets have emerged as a promising alternative for optimizing Image Diffusion models \citep{gflownet_diffusion, gflownet_diffusion_toy, db_on_images}. In this work, we introduce AbFlowNet to address these concerns in biophysical energy optimization. As shown in Figure \ref{fig:openfig}, we reframe the denoising diffusion process \citep{ho2020denoising, leach2022denoising} in a GFlowNet \citep{gflownet1} framework where each partially denoised structure is a GFlowNet \textit{state} and the forward and backwards transition probabilities are \textit{flows through edges}. The \textit{flow of a state} is the sum of all flows of all trajectories through that state. The final fully-denoised state's flow is the binding energy \textit{reward}. We use the Trajectory Balance objective \citep{gflownet1} to enforce that the forward and backward flow for a trajectory (a full denoising sequence starting from random noise and ending at a CDR structure) must be equal. As a result, the diffusion model implicitly learns better state transitions that lead to higher rewards.

Practically, AbFlowNet is a training-time recipe applicable to any diffusion backbone: a single learned parameter and a TB loss term added to the existing diffusion losses. We instantiate AbFlowNet on DiffAb \citep{diffab}, a widely used open-source diffusion backbone with publicly available training code, which enables a controlled and reproducible evaluation of the proposed objective. Under this setup, AbFlowNet consistently improves over the diffusion-only baseline for the same number of training epochs. Concretely, when averaged over all six CDR regions, AbFlowNet improves amino acid recovery (AAR) by $3.06\%$, root mean square deviation (RMSD) by $20.40\%$, and generates $3.60\%$ more CDRs with higher binding energy than the reference CDR. AbFlowNet also improves over DiffAb in Top-1 total energy and binding energy by $24.8\%$ and $38.1\%$. Unlike online RL approaches such as AbDPO \citep{abdpo}, AbFlowNet is orders of magnitude less expensive, does not need repeated use of unreliable energy estimators, and does not rely on pseudo-labeling the test set. Our key contributions are:
\begin{enumerate}

    \item We present AbFlowNet, the first application of the GFlowNet framework for direct binding energy optimization in \textit{de novo} diffusion-based CDR design. AbFlowNet improves over the base diffusion model in all metrics.
    \item AbFlowNet is competitive with RL-based methods \citep{abdpo} without using the test set complexes to generate synthetic CDR data for training, thereby mitigating data leakage concerns.
    \item Unlike existing RL-based approaches \citep{abdpo, alignab} which reduce AAR and RMSD, AbFlowNet improves AAR by $+3.06\%$ and RMSD by $+20.40\%$.
    \item AbFlowNet's gains transfer to an orthogonal energy estimator and preserve generation diversity, addressing two known failure modes of energy-optimized generative models.

\end{enumerate}

\section{Related Works}

\paragraph{Computational CDR Design}
Classical approaches to CDR design, such as RAbD ~\cite{adolf2018rosettaantibodydesign} and AbDesign ~\cite{lapidoth2015abdesign}, rely on Monte Carlo algorithms that sample and optimize antibody structures based on biophysical energy functions. These methods, while effective in certain contexts, are computationally expensive and often get trapped in local optima due to the rugged energy landscape \citep{diffab, dymean}. In recent years, deep learning methods have emerged as promising alternatives. Notable Graph Neural Network-based models include HERN \citep{hern}, MEAN \cite{mean}, AbGNN \citep{abgnn}, and dyMEAN \citep{dymean}. These methods have shown high AAR and RMSD but are limited in generation diversity due to their GNN structure. Diffusion-based models can generate multiple CDRs given a complex, which can be later ranked heuristically. Notable diffusion-based co-design models include AbDiffuser \citep{martinkus2024abdiffuser}, DiffAb \citep{diffab}, AbDesign \citep{abdesign_abdock}, AbX \citep{abx}, and IgGM \citep{iggm}; AbX and IgGM use large Protein Language Models \citep{esm2} to enforce evolutionary plausibility. AbFlow \citep{wang2026abflow} proposed a flow-matching based architecture for CDR design. A complementary line of work generates antibody backbones first and fills in sequences via inverse folding, including RFDiffusion-Antibody \citep{rfdiffusion_antibody} and Germinal \citep{germinal} which use inverse-folding models such as ProteinMPNN \citep{proteinmpnn}, AbMPNN\citep{abmpnn} and AntibodyDesignBFN \citep{hu2026antibodydesignbfn}. AbFlowNet is a training-time modification of any diffusion co-design model and is therefore orthogonal to architectural advances in this line of work.
\vspace{-1em}
\paragraph{Binding Energy Optimization For CDR Design} A key metric for CDR effectiveness is binding affinity. One commonly used energy metric is binding energy $ \Delta G$. Since binding energy is a singular value for the entire complex, it is a sparse training signal which is often optimized via Reinforcement Learning \citep{sutton1998reinforcement}. AbDPO \citep{abdpo} post-trains a base DiffAb model by repeatedly sampling new CDRs, ranking them based on binding affinity, determined with \texttt{Rossetta InterfaceAnalyzer} \citep{chaudhury2010pyrosetta} and using DPO \citep{dpo}. However, this RL training phase significantly lowers AAR and RMSD compared to the base method. AlignAb \citep{alignab} points out that there are multiple valid energy-based rewards and finetune separate models for each reward using DPO. AbNovo \citep{abnovo} follows the approach of AbDPO with AbX instead of DiffAb as the base model and uses Noise Contrastive Alignment (NCA) \citep{nca_rl} as the RL objective instead of DPO. One notable weakness of all online RL methods is the need to compute binding energy for newly designed CDRs. \textit{In silico} methods such as \texttt{Rosetta} \citep{chaudhury2010pyrosetta, adolf2018rosettaantibodydesign} or \texttt{OpenMM Yank} \citep{eastman2017openmm, yank2020} have only moderate corelation with the real binding energy\citep{binding_energy_survey}. Furthermore, the generated CDRs are not guaranteed to be geometrically plausible, which might reduce the reliability of energy estimators further. In contrast, AbFlowNet does not require computing the energy of newly generated CDRs and can, in principle, be trained solely on \textit{in vitro} affinity data of CDRs in the training set.

\section{Background}
\subsection{Antibody-Antigen Complex}

\begin{wrapfigure}{l}{0.5\linewidth}

    \centering
    \includegraphics[width=1\linewidth]{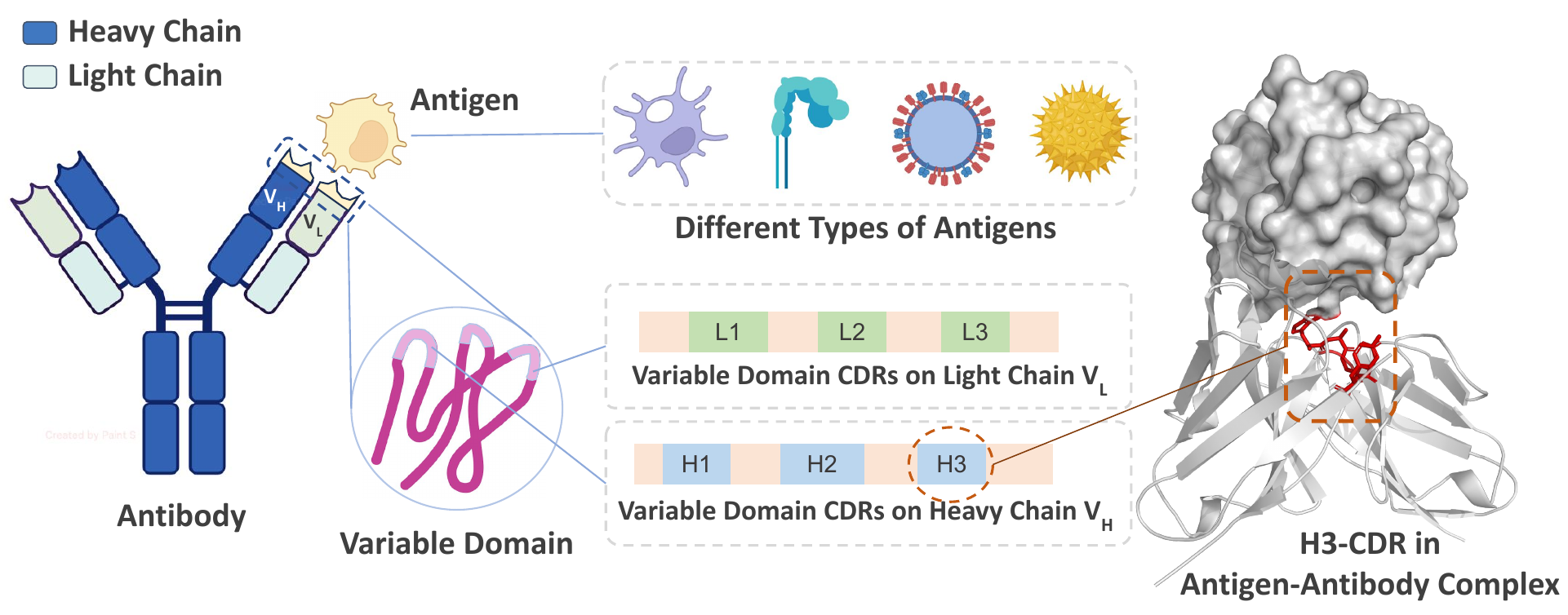}
    \caption{\footnotesize Overview of antibody-antigen complex structure. Left: Y-shaped antibody composed of heavy (V\_H) and light (V\_L) chains. Center: the variable domain contains six CDR loops, which determine antigen recognition. Right: structural view of the complex with the CDR-H3 loop highlighted.}
    \label{fig:aa_complex}

\end{wrapfigure}

As shown in Figure~\ref{fig:aa_complex}, each antibody variable domain contains three complementarity-determining regions (CDRs), giving six CDRs per antibody in total. These hypervariable loops---shaped by genetic recombination and somatic hypermutation---form a binding site complementary in geometry and chemistry to the antigen epitope, interacting through non-covalent forces such as hydrogen bonds, electrostatic interactions, and van der Waals contacts \citep{cdr_basics}. Among the six CDRs, CDR-H3 is particularly consequential: its exceptional sequence diversity makes it the primary determinant of binding specificity and affinity. In our work, we aim to design CDR sequences and structures conditioned on the antibody's non-CDR scaffold and the target antigen.

\subsection{GFlowNet}

GFlowNets \citep{gflownet1} learn to sample from a positive reward $R: \mathcal{X} \rightarrow \mathbb{R}_{\geq 0}$ over the terminal states $\mathcal{X}$ of a DAG $G = (\mathcal{S}, \mathcal{A})$, with sampling probability proportional to the reward. A GFlowNet defines a flow $F$ on trajectories $\tau = (s_0 \rightarrow s_1 \rightarrow ... \rightarrow s_n)$, inducing the \textbf{\textit{state flow}} $F(s)=\sum_{\tau \ni s}F(\tau)$, the \textbf{\textit{edge flow}} $F(s\rightarrow s')=\sum_{\tau \ni (s\rightarrow s')}F(\tau)$, the terminal-state condition $F(s_n)= R(s_n)$, and forward/backward transition policies $P_F(s'|s) = F(s \rightarrow s')/F(s)$, $P_B(s|s') = F(s \rightarrow s')/F(s')$.

\paragraph{Trajectory Balance} Trajectory Balance (TB) \citep{trajectory_balance} enforces consistency between forward generation and backward reconstruction over an entire trajectory $\tau = (s_0 \rightarrow s_1 \rightarrow ... \rightarrow s_n)$. Since the initial-state flow $F(s_0)=\sum_\tau F(\tau)$ is intractable, it is approximated by a learned scalar $Z_\theta$, yielding the constraint:
\vspace{-0.5em}
\begin{equation}
Z_\theta\prod_{t=1}^n P_F(s_t|s_{t-1}) = R(s_n)\prod_{t=1}^n P_B(s_{t-1}|s_t)
\end{equation}

\section{Methodology}

In Section \ref{method_denoising}, we discuss the training objectives of the base diffusion model, and in Section \ref{method_tb}, we describe our reframing of the diffusion process as a GFlowNet. Equation \ref{eqn:loss_combo} presents our final training objective, which jointly optimizes the diffusion losses and the binding energy.

\subsection{Denoising Objective}
\label{method_denoising}
We train a diffusion probabilistic model parameterized by a neural network for CDR design. Following \citet{diffab}, we condition on the antigen structure and the antibody framework to generate CDR. The model is trained using the standard denoising objective across three protein properties: \textbf{amino acid type} \(d_i \in \{A,C,D,E,F,G,H,I,K,L,M,N,P,Q,R,S,T,W,Y,V\}\), \textbf{3D coordinate} \(x_i \in \mathbb{R}^3\), and \textbf{3D orientation}  \(O_i \in SO(3)\) where $SO(3)$ is the Lie group of 3D rotations.

Assume the CDR to be generated has \(m\) amino acids with index from \(l+1\) to \(l+m\). They are denoted as $S^t =
 \{ s^t_j | j = l+1, \ldots, l+m \}$ where $s^t_j = (d^t_j, x^t_j, O^0_j)$. Our goal is to model the distribution of \(S^0\) given the structure of the antibody-antigen complex \(C = \{(d_i, x_i, O_i)\ |\ i \in \{1, \ldots, N\} \setminus \{l+1, \ldots, l+m\}\}\).

\paragraph{Multinomial Diffusion for Amino Acid Types}
The forward diffusion process for amino acid types is based on the multinomial diffusion process \citep{hoogeboom2021argmax}:

\vspace{-1.5em}
\begin{equation}
q(d_j^t | d_j^{t-1}) = \text{Mul} \left( (1 - \beta_{\text{type}}^t) \cdot \text{oh}(d_j^{t-1}) +   \frac{\beta_{\text{type}}^t}{20} \cdot \mathbf{1} \right)
\end{equation}

\vspace{-1.0em}

\begin{equation}
\label{eqn:AA_back}
p_\theta(d^{t-1}_j | S^t, C) = \text{Mul}\left(F_\theta(S^t, C)[j]\right)
\end{equation}
where $q$ is the forward diffusion process, $p$ is the backward denoising process, $\text{Mul}()$ is the Multinomial function and $\text{oh}()$ is the one-hot function. \(\beta^t_{\text{type}}\) is the probability of resampling another amino acid uniformly over the 20 types, and \(F_\theta(\cdot)[j]\) is a neural network model that predicts the probability of the amino acid type for the \(j\)-th amino acid on the CDR. The training objective is to minimize the expected KL divergence between the posterior distribution $q$ and the predicted distribution $p$:

\begin{equation}
    L^t_{\text{type}} = \mathbb{E}\left[\frac{1}{m} \sum_{j} D_{\text{KL}} \left(q(d^{t-1}_j | d^t_j) \| p(d^{t-1}_j | S^t, C)\right)\right]
\end{equation}
\vspace{-1.0em}
\paragraph{Diffusion for 3D Coordinates}

The forward and backward diffusion processes for the coordinate \(x_j\) are defined as:
\vspace{-0.5em}
\begin{equation}
q(x^t_j | x^{t-1}_j) = \mathcal{N}\left(x^t_j \bigg| \sqrt{1 - \beta^t_{\text{pos}}} \cdot x^{t-1}_j, \beta^t_{\text{pos}} I \right)
\end{equation}
\vspace{-0.5em}
\begin{equation}
p_\theta(x^{t-1}_j | S^t, C) = \mathcal{N}\left(x^{t-1}_j \bigg| \mu_{p_\theta}(S^t, C), \beta^t_{\text{pos}} I \right)
\end{equation}
\vspace{-0.5em}
\begin{equation}
\label{eqn:pos_back}
\mu_{p_\theta}(S^t, C) = \frac{1}{\sqrt{1-\beta^t_{\text{pos}}}} \left(x^t_j - \frac{\beta^t_{\text{pos}}}{\sqrt{1 - \tilde{\alpha}^t_{\text{pos}}}} G_\theta(S^t, C)[j] \right)
\end{equation}

where \(\beta^t_{\text{pos}}\) controls the rate of diffusion in $q$. The denoising diffusion process $p$ uses the reparameterization trick \citep{ho2020denoising} and \(G_\theta(\cdot)[j]\) is a neural network that predicts the standard Gaussian noise \(\epsilon_j \sim \mathcal{N}(0, I)\) added instead of predicting $x^{t-1}_j$ directly where \(\tilde{\alpha}^t_{\text{pos}} = \prod_{\tau=1}^{t} (1 - \beta^\tau_{\text{pos}})\). The training objective is to minimize the expected MSE between \(G\) and \(\epsilon\):

\vspace{-1.0em}
\begin{equation}
L^t_{\text{pos}} = \mathbb{E}\left[\frac{1}{m} \sum_j \| \epsilon_j - G_\theta(S^t, C)[j] \|^2\right]
\end{equation}

\paragraph{SO(3) Denoising for Amino Acid Orientations}
Following \citet{leach2022denoising, diffab}, the denoising process for orientation directly attempts to predict the final orientation $O^0_j$ from  $O^t_j$. The transitions are defined as:

\begin{equation}
\label{eqn:rot_forward}
q(O^t_j | O^{0}_j) = \mathcal{IG}_{\text{SO(3)}}\!\Bigl(O^t_j | \lambda\!\bigl(\sqrt{\bar{\alpha}^t_{ori}},\,O^{0}_j\bigr),\;1-\bar{\alpha}^t_{ori}\Bigr)
\end{equation}
\vspace{-0.5em}
\begin{equation}
\label{eqn:rot_back}
p_\theta(O^{t-1}_j | S^t, C) = \mathcal{IG}_{\text{SO(3)}}\left(O^{t-1}_j \bigg| H_\theta(S^t, C)[j], \beta^t_{\text{ori}}\right)
\end{equation}

where $\bar{\alpha}^t_{ori}=\prod_{\tau=1}^t(1-\beta_{ori}^\tau)$ with $\beta_{ori}^t$ being the variance increased with step $t$, $\mathcal{IG}_{\text{SO(3)}}$ denotes the isotropic Gaussian distribution on \(SO(3)\) \citep{leach2022denoising} and $\lambda(\gamma, x) \;=\; \exp\!\bigl(\gamma\,\log(x)\bigr)$ is the geodesic interpolation (or “scaling”) of the rotation \(x \in SO(3)\) from the identity. \(H_\theta(\cdot)[j]\) is a neural network that denoises the orientation and outputs the denoised orientation matrix. The training objective simply minimizes the difference between the real and predicted orientation matrices:
\vspace{-0.5em}
\begin{equation}
L^t_{\text{ori}} = \mathbb{E}\left[\frac{1}{m} \sum_j \| (O^0_j)^\top \tilde{O}^{t-1}_j - I \|^2\right]
\end{equation}
where $\tilde{O}^{t-1}_j = H_\theta(S^t, C)[j]$ is the predicted denoised orientation. Since $\tilde{O}^{t-1}_j$ directly tries to predict ${O}^{0}_j$ instead of ${O}^{t}_j$, the formulation does not strictly follow the framework of diffusion probabilistic models, though the overall principle is the same.

\subsection{Trajectory Balance Objective}
\label{method_tb}
In addition to the denoising objectives, we aim to optimize the binding energy of the generated CDR with respect to the antigen and antibody. However, binding energy can only be computed for the final CDR after a complete denoising process, making it a sparse reward. To address this, we use the \textbf{Trajectory Balance (TB)} objective \citep{trajectory_balance} which propagates rewards back through the diffusion trajectory by enforcing global flow-matching constraints.

We begin by defining a GFlowNet state as the partially denoised CDR at timestep $t$. A CDR is composed of a sequence of amino acids, and the transition probability for each amino acid location $j$ is the product of the three independent denoising processes (Equations \ref{eqn:amino_acid_q} and \ref{eqn:amino_acid_p}). We define the GFlowNet transition probability of the entire CDR as the product of the probability of each location (Equations \ref{eqn:CDR_q} and \ref{eqn:CDR_p}):

\vspace{-0.5em}
\begin{equation}
\label{eqn:amino_acid_q}
q(s^{t}_j|s^{t-1}_j)= q(d^{t}_j | d^{t-1}_j) \cdot  q(x^{t}_j | x^{t-1}_j) \cdot  q(O^{t}_j | O^{0}_j)
\end{equation}

\vspace{-0.5em}
\begin{multline}
\label{eqn:amino_acid_p}
p_\theta(s^{t-1}_j|s^{t}_j, C) = p_\theta(s^{t-1}_j|S^t, C)
= p_\theta(d^{t-1}_j | S^t, C) \cdot p_\theta(x^{t-1}_j | S^t, C) \cdot p_\theta(O^{t-1}_j | S^t, C)
\end{multline}

\begin{minipage}{0.45\linewidth}
\begin{equation}
\label{eqn:CDR_q}
q(S^{t}|S^{t-1})= \prod_{j=l+1}^{l+m} q(s^{t}_j|s^{t-1}_j)
\end{equation}
\end{minipage}
\hfill
\begin{minipage}{0.45\linewidth}
\begin{equation}
\label{eqn:CDR_p}
p_\theta(S^{t-1}|S^{t}, C)= \prod_{j=l+1}^{l+m} p_\theta(s^{t-1}_j|s^{t}_j, C)
\end{equation}
\end{minipage}

For each data point during a mini-batch update, we uniformly sample a timestep $t$ to compute $L^t_{\text{type}}$, $L^t_{\text{pos}}$, and $L^t_{\text{ori}}$. However, we require a complete trajectory to apply TB. Therefore, we compute all denoising probabilities $p_\theta(S^{t-1}|S^t, C)$ and noising probabilities $q(S^{t}|S^{t-1})$ for $t \in \{1, \dots, T\}$. Following \citet{kim2024ant}, we precompute the reward $\mathcal{R}(S^0)=\exp(-\alpha\cdot\text{BindingEnergy}(S^0))$ for each CDR $S^0$ in the training dataset. Since the binding energy is typically negative for real CDRs and lies roughly in the range $[-100, 0]$, we set $\alpha=10^{-2}$, yielding rewards in the range $[1, e]$. Positive rewards are a deliberate choice, as all complexes in the training dataset were experimentally determined and thus exhibit baseline energy stability. We enforce the TB objective:
\vspace{-0.5em}
\begin{equation}
L_{\text{TB}} =\left(\log\frac{Z_\theta\prod_{t=1}^T p_\theta(S^{t-1}|S^{t}, C)}{\mathcal{R}(S^0)\prod_{t=1}^T q(S^{t}|S^{t-1})}\right)^2
\label{eqn:TB_loss}
\end{equation}

where, following \citet{trajectory_balance}, \(Z_\theta\) is the model’s estimate of the global partition function (normalizing constant over all terminal states) and is therefore a single scalar for the whole distribution, and \(\mathcal{R}(S^0)\) is the binding energy reward. Therefore, the overall training objective combines the denoising losses and the TB objective:
\begin{equation}
L = \mathbb{E}_{t \sim \text{Uniform}(1,\ldots,T)}\left[L^t_{\text{type}} + L^t_{\text{pos}} + L^t_{\text{ori}} \right]+w\cdot L_{\text{TB}}
\label{eqn:loss_combo}
\end{equation}

where $w$ is a scaling factor for balancing the denoising objectives computed per diffusion step and the TB objective calculated over the entire trajectory. The TB term thus acts as a reward-informed regularizer on the denoiser alongside the reconstruction losses.

\subsection{Sampling Algorithm}
\label{method_sampling}

For sequence-structure co-design, we construct $S^T$ by sampling amino acid types for each position from the distribution  $d_j^T \sim \text{Uniform}(20)$, CDR positions from the standard normal distribution: $x_j^T \sim \mathcal{N}(0, I_3)$, and orientations from the uniform distribution over SO(3): $O_j^T \sim \text{Uniform(SO(3))}$. AbFlowNet iteratively denoises the sequence and structures following the standard diffusion process until $t=0$.  Upon generating the amino acid sequence and the structure of the backbone, we optimize the side-chain angles using PyRosetta \texttt{PackRotamersMover} \citep{chaudhury2010pyrosetta}.

Crucially, AbFlowNet applies the GFlowNet balance objective solely during training, not during inference. This approach enables AbFlowNet to operate as a standard diffusion model at sampling time, without requiring energy calculations via the \texttt{Rosetta InterfaceAnalyzer}.

\section{Experiments}
\label{experiments}

\paragraph{Dataset Curation}
We use the Structural Antibody Database (SAbDab) \citep{dunbar2014sabdab} as the training dataset.  We first remove structures whose resolution is less than 4\AA\ and discard antibodies targeting non-protein antigens \citep{diffab}. Our final training dataset contains 9410 antigen-antibody complexes. We evaluate sequence-structure codesign on the RAbD test dataset, consisting of 60 diverse antibody-antigen complexes \cite{rabd}. We also evaluate on the test set proposed by DiffAb \citep{diffab}, which contains 19 complexes with antigens from several well-known pathogens, including SARS-CoV-2, MERS, influenza, and so on.  For both test sets, we strictly remove the overlap between the training set and the testing sets using a CDR-H3 sequence identity threshold of 50\%.

\paragraph{Model Architecture and Hyperparameters}
We use the transformer-based parametrization of \citet{diffab} to encode the antigen-antibody complex $C$ and follow \citet{trajectory_balance} in parameterizing the initial state flow $Z$ as a single learnable scalar. Both DiffAb and AbFlowNet are trained for 200K Adam \citep{kingma2014adam} steps at learning rate $10^{-6}$: 195K diffusion-only steps followed by 5K steps with the fused diffusion+TB loss at weight $w = 5\times10^{-6}$. We use $N=100$ samples per complex in Table~\ref{tab:energy_h3} for efficiency; full sweep over $N$ and $w$ are in Appendices~\ref{app:sampling_sweep} and \ref{app:ablate_w}.

\paragraph{Metrics} We use standard metrics to evaluate designed antibodies \citep{rabd, diffab, abx}, namely, (1) \textbf{AAR}: the amino acid recovery rate measured by the sequence identity between the reference CDR sequences and the generated sequences, (2) \textbf{RMSD}: the $C_\alpha$ root-mean-square deviation (RMSD) between the generated structure and the original structure, and (3) \textbf{IMP}: the percentage of designed CDRs with lower (better) binding energy ($\Delta G$) than the original CDR. The binding energy is calculated by \texttt{InterfaceAnalyzer} in the Rosetta software package \citep{adolf2018rosettaantibodydesign, chaudhury2010pyrosetta}. Diffusion models are capable of generating diverse data points from the target distribution by randomly sampling from the initial distribution \citep{ho2020denoising, leach2022denoising}. This is especially advantageous in CDR design, where we can generate multiple candidate CDRs \textit{in silico} and select only the most promising CDR for \textit{in vitro} validation according to some desirable property. To this end, the authors of AbDPO \citep{abdpo} proposed: (1) \textbf{Top-1 CDR} $\mathbf{E_{total}}$: total energy of the whole designed CDR (kcal/mol) of the \textit{best CDR} out of $N$; (2) \textbf{Top-1 CDR-Ag} $\mathbf{\Delta G}$: the difference in total energy between the bound state and the unbound state of that CDR and antigen. Following AbDPO, we generate $N$ CDRs for each complex in the RAbD test dataset and choose the best CDR ranked by $\mathbf{E_{total}}+\mathbf{\Delta G}$.

\section{Results}

\begin{table*}[!h]
\centering
\caption{
\textbf{Top-1} CDR $E_{\text{total}}$ and CDR-Ag $\Delta G$ (kcal/mol) for \textit{de novo} CDR-H3 design. $(\downarrow)$ indicates lower is better. We also show the percentage reduction over DiffAb with the same sampling budget. AbX and AbNovo report Mean CDR $E_{\text{total}}$ and CDR-Ag $\Delta G$ instead of Top-1.
}
\resizebox{0.99\textwidth}{!}{%
\begin{tabular}{l|c|cc|c|c}
\toprule
\textbf{Methods} & \textbf{Sampling} & \textbf{CDR $E_{\text{total}}$ (↓)} & \textbf{CDR-Ag $\Delta G$ (↓)} & \textbf{Test Set Used} & \textbf{\# Param. (M)} \\
 & \textbf{Budget} & (with $\Delta$) & (with $\Delta$) & \textbf{in Sampling} & \textbf{} \\
\midrule
Reference && 4.52 & -13.72 & — &  —  \\
\midrule
\multicolumn{6}{l}{\textbf{GNN Baselines}} \\
\midrule
HERN \citep{hern} & 2,528 & 7594.94 & 1159.34 & — & 5 \\
MEAN \citep{mean} & 2,528 & 3113.70 & 114.98 & — & - \\
dyMEAN$^1$ \citep{dymean} & 2,528 & 15025.67 & 2391.00 & — & - \\
dyMEAN$^2$ & 2,528 & 3234.30 & 1619.24 & — & - \\
\midrule
\multicolumn{6}{l}{\textbf{Diffusion-Based}} \\
\midrule
DiffAb \citep{diffab} & 2,528 & 211.00 & 9.54 &  —  & 4 \\
AbDPO \citep{abdpo}$^\dagger$ & 2,528 &
162.75 (↓23.4\%)& \textbf{-4.85 (↓61.9\%)} & Yes & 4 \\
\midrule
DiffAb & 100 & 480.25 & 11.20 &  —  & 4 \\
AbFlowNet (Ours) & 100 & \textbf{362.03 (↓24.8\%)} & 1.71 (↓38.1\%) & \textbf{No} & 4 \\
\midrule
Reference &  &  —  & -19.41 & — &  —  \\
DiffAb & 128 & —  & -0.96  & No & 4\\
AbX\citep{abx}    & 128 & —  & 4.79 (↑31.2\%)  & No & 12 (+ ESM-3B)\\
AbNovo\citep{abnovo} & 128 & —  & -12.05 (↓60.1\%)& No & 12 (+ ESM-3B)\\
\bottomrule
\end{tabular}
}
\begin{flushleft}
\footnotesize{$^\dagger$AbDPO is not open-sourced and cannot be independently reproduced; numbers are reproduced from \citet{abdpo}. Improvements for AbDPO and AbFlowNet are relative to DiffAb under the same sampling budget.}
\end{flushleft}
\label{tab:energy_h3}
\vspace{-1em}
\end{table*}

\subsection{AbFlowNet Greatly Improves Binding Energy of generated CDR-H3}
\label{ssec:top_1}

\begin{wrapfigure}{r}{0.5\linewidth}
\centering
\includegraphics[width=1\linewidth]{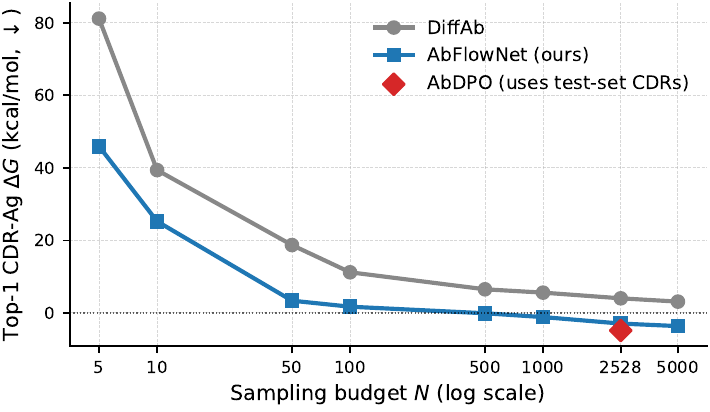}
\caption{Top-1 CDR-Ag $\Delta G$ on the RAbD test set as a function of sampling budget $N$. AbDPO's reported value ($-4.85$ at $N=2{,}528$) is shown for reference. Exact values are reported in Appendix~\ref{app:sampling_sweep}}
\label{fig:sampling_sweep}
\end{wrapfigure}

Table~\ref{tab:energy_h3} shows that on \textit{de novo} CDR-H3 generation, AbFlowNet greatly improves the binding energy over the base diffusion model and is competitive with AbDPO (without relying on test-set structures. The H3 region is especially difficult for all models to generate because the H3 loop undergoes independent mutation before joining the rest of the antibody sequence \citep{graves2020review}, introducing variability and significantly affecting the structure and function of the antibody. Specifically, AbFlowNet significantly outperforms DiffAb at $N = 100$, achieving performance gains comparable to AbDPO~\citep{abdpo} without large-scale synthetic CDR generation and energy estimation. To evaluate relative improvement, we apply the formula $\Delta = \textsc{(\text{method} - \text{baseline}) / (\text{reference} - \text{baseline})}$, which normalizes performance gains with respect to the baseline and reference. We further validate our findings on the DiffAb test benchmark~\citep{diffab}, which includes 19 complexes with antigens from SARS-CoV-2, MERS, and influenza; full results are provided in Appendix Table~\ref{tab:diffab_energy_h3}. We verify that the CDRs generated by AbFlowNet remain evolutionarily plausible using ESM-2 \citep{esm2} in Appendix \ref{app:evoplaus} and find that the ESM embeddings of generated CDRs strongly cluster with their reference.

Compared to AbDPO~\citep{abdpo}, which is an online RL method that post-trains a DiffAb model to optimize CDR binding energy, AbFlowNet offers several advantages. AbDPO samples 10,122 CDRs per test complex to construct a preference dataset, and updates the model via Direct Preference Optimization (DPO)~\citep{dpo}. This approach has two main limitations: (1) sampling at this scale and computing binding energies is computationally expensive, and (2) using test-set antibody–antigen complexes during RL introduces potential bias. In contrast, AbFlowNet relies solely on precomputed binding energies from the 9,410 training examples, eliminating the need for expensive sampling and reward computation during optimization.

Finally, we note that both CDR $E_{\text{total}}$ and CDR-Ag $\Delta G$ are \textit{Top-1} metrics that select the best-scoring sample among $N$ generated CDR-H3s. These metrics are inherently sensitive to the sampling budget. We report distribution plots and statistics for individual complexes in Appendix \ref{app:distribution}. DiffAb’s $E_{\text{total}}$ improves significantly from 480.25 kcal/mol at $N = 100$ to 211.00 kcal/mol at $N = 2,528$. Figure \ref{fig:sampling_sweep} shows that AbFlowNet uniformly outperforms DiffAb across budgets, with the largest gains at small $N$ where the metric is most relevant for compute-constrained \textit{in silico} screening. At $N=2{,}528$, AbFlowNet achieves $\Delta G = -2.94 $ against AbDPO's reported value of $-4.85$. We note that AbFlowNet was not optimized against the test set complexes in any capacity, while AbDPO specifically performs DPO optimization on RAbD test complexes.

\subsection{AbFlowNet's Gains Transfer to FoldX, an Independent Energy Estimator}
\label{ssec:orthogonal}
\vspace{-0.5em}

The PyRosetta-derived training reward is an imperfect proxy for true binding affinity \citep{binding_energy_survey}, raising the concern that AbFlowNet's gains over DiffAb might be specific to the PyRosetta scoring function. To check that the gains transfer, we re-score every generated CDR with \texttt{FoldX} \citep{foldx}, an independent biophysical force field used in clinical antibody design workflows \citep{desautels_foldx}.

\begin{wraptable}{r}{0.33\linewidth}
\vspace{-1.2em}
\centering
\caption{FoldX $\Delta G$ on RAbD ($\downarrow$, kcal/mol).}
\label{tab:foldx_main}
\small
\begin{tabular}{l|cc}
\toprule
\textbf{CDR} & \textbf{DiffAb} & \textbf{AbFlowNet} \\
\midrule
H1 & -8.000 & \textbf{-9.067} \\
H2 & -8.399 & \textbf{-9.605} \\
H3 & -2.112 & \textbf{-3.246} \\
L1 & -8.463 & \textbf{-9.295} \\
L2 & -8.505 & \textbf{-9.542} \\
L3 & -7.058 & \textbf{-7.934} \\
\bottomrule
\end{tabular}
\vspace{-1em}
\end{wraptable}

Table~\ref{tab:foldx_main} shows that AbFlowNet improves over DiffAb on \emph{all six} CDRs under FoldX, confirming the gains are not an artifact of optimizing a single proxy. A paired comparison on CDR-H3 at an equal budget of 100 candidates per target confirms this is not driven by a few outliers: AbFlowNet attains a lower Top-1 FoldX energy on 44 of 60 RAbD complexes, with a mean paired improvement of $2.13$ kcal/mol (95\% paired-bootstrap CI $[0.78, 3.66]$) and a two-sided Wilcoxon signed-rank test giving $p = 2.76 \times 10^{-5}$. On the head-to-head AbX (CDR-H3) comparison with public weights, AbX achieves $-5.13$ kcal/mol using an 11.86M-parameter backbone with a frozen ESM-3B encoder, vs.\ AbFlowNet, which achieves $-3.246$ kcal/mol using the 4M-parameter DiffAb backbone --- consistent with our framing that AbFlowNet contributes a training objective orthogonal to architectural advances (see §\ref{sec:discussion}). We further validate generalization on a \textbf{novel-antigen subset of RAbD}, where AbFlowNet reduces mean CDR-Ag $\Delta G$ from 25.47 to 10.03 kcal/mol (Appendix~\ref{app:novel_antigen}).

\subsection{Joint Optimization Outperforms Diffusion-only Baseline}
\label{ssec:aar_rmsd}
\vspace{-0.5em}
While CDR $E_{\text{total}}$ and CDR-Ag $\Delta G$ measure only binding energy, we evaluate three metrics that compare the generated CDRs to the reference CDR in the dataset. We find that AbFlowNet outperforms DiffAb in all three metrics: $+3.06\%$ in AAR, $+20.40\%$ in RMSD, and $+3.60\%$ in IMP when averaged over all 6 CDR regions. We include detailed results in Appendix \ref{rabd_results}. For the CDR-L3 chain, in particular, AbFlowNet achieves an RMSD considerably lower than that of other methods.  This shows that our Diffusion-GFlowNet Joint Optimization framework maintains strong reconstruction performance, whereas DPO-based post-training RL reduces it. AbDPO, which post-trains only on binding energy, does not preserve the training distribution as well and leads to a 9.96\% reduction in average AAR and an increase in RMSD by 0.14~\AA. Therefore, AbFlowNet improves binding energy without sacrificing structural accuracy. We find consistent improvements in most CDR regions when using the test set proposed by DiffAb \citep{diffab}, shown in Appendix Table \ref{tab:diffab_test_results}.

\paragraph{Diversity preservation.} Reward-driven training risks mode collapse onto a single high-reward CDR. The mean pairwise sequence identity among 100 generations per complex is essentially unchanged from DiffAb (54.46\% vs.\ 52.63\%; Appendix~\ref{app:diversity}). This is consistent with the GFlowNet objective: TB enforces probability $\propto$ reward rather than maximization \citep{gflownet1}, so concentrating mass on a single mode would itself violate the loss. RL/DPO methods do not have this property.

\subsection{AbFlowNet Designs CDRs with Tighter Antigen Contact}
\label{ssec:qual}
\vspace{-0.5em}

\begin{figure*}[h]
  \centering
  \begin{subfigure}{0.24\textwidth}
    \includegraphics[width=\linewidth]{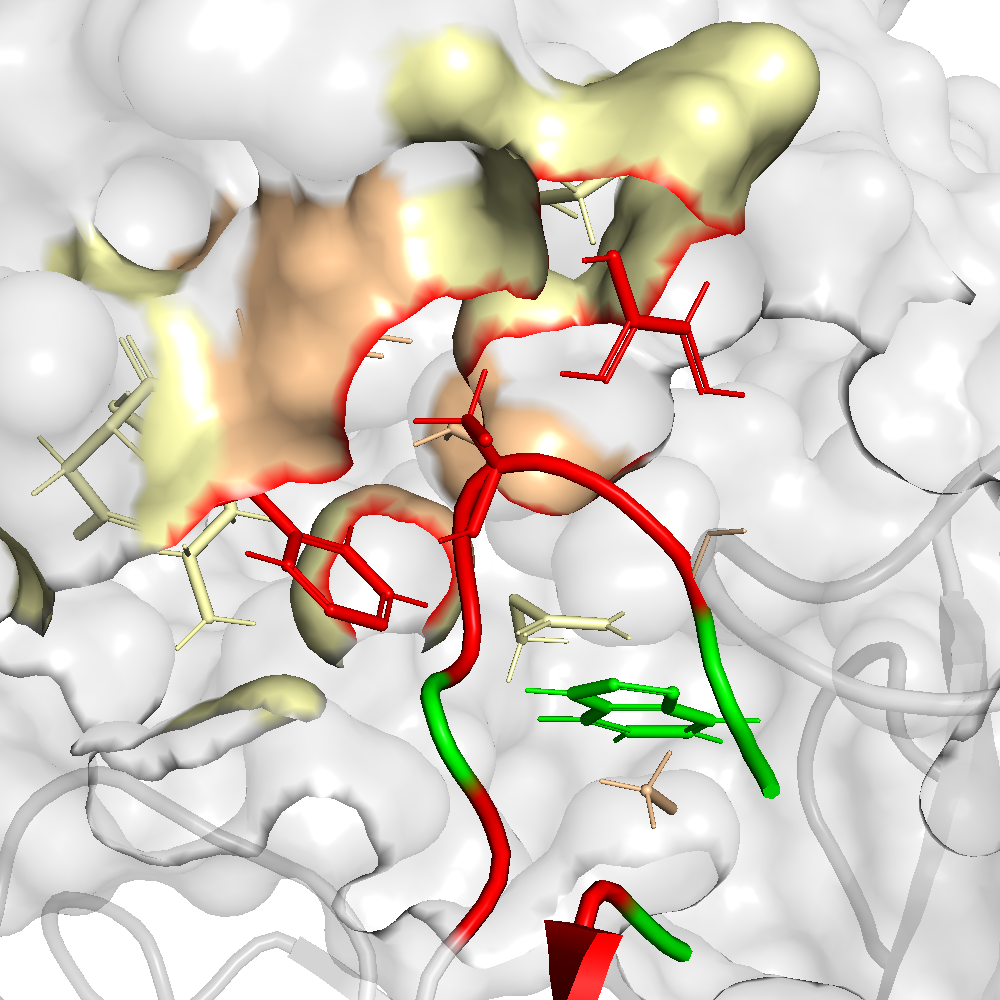}\captionsetup{justification=centering}
    \caption{DiffAb H3 \\$E_{total} =18.59 \text{ kcal/mol}$ \\$\Delta G=-2.52 \text{ kcal/mol}$}
  \end{subfigure}
  \hfill
  \begin{subfigure}{0.24\textwidth}
    \includegraphics[width=\linewidth]{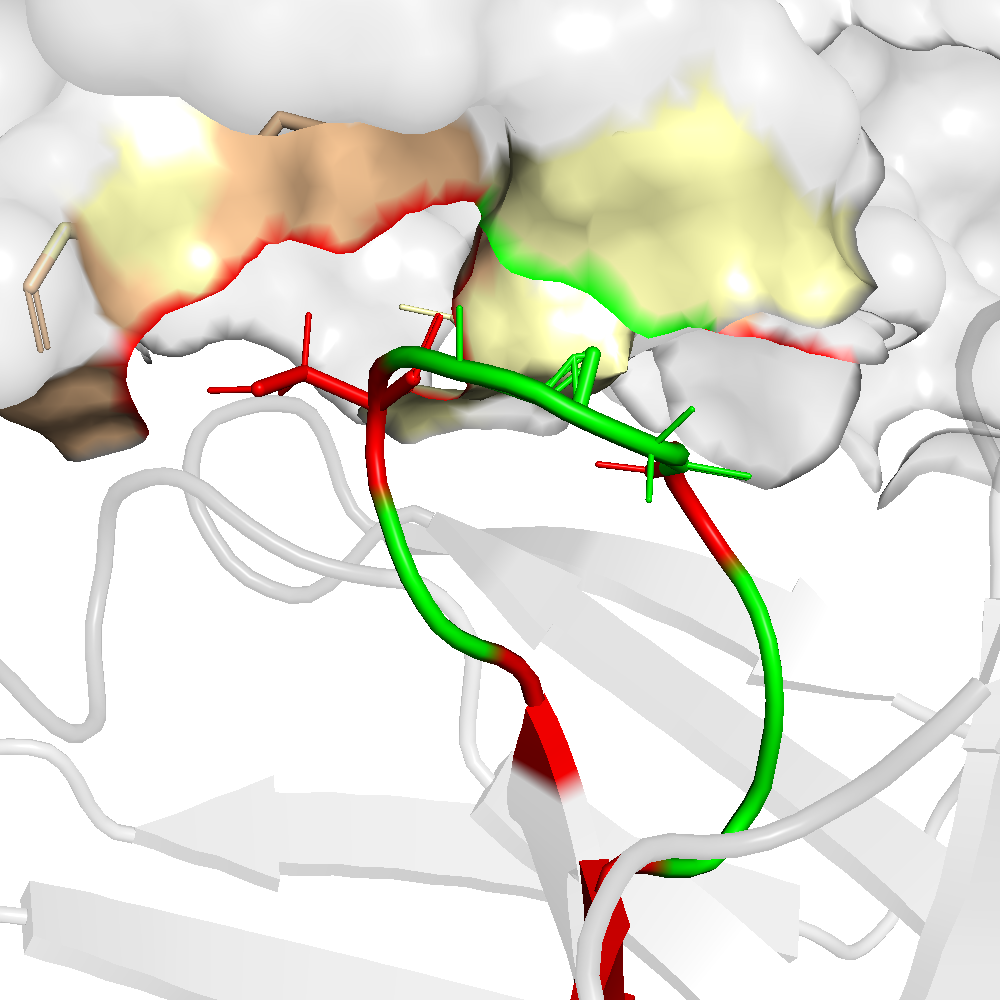}     \captionsetup{justification=centering}
    \caption{AbFlowNet H3 \\$E_{total} =-8.60 \text{ kcal/mol}$ \\$\Delta G=-19.79 \text{ kcal/mol}$}
  \end{subfigure}
  \hfill
  \begin{subfigure}{0.24\textwidth}
    \includegraphics[width=\linewidth]{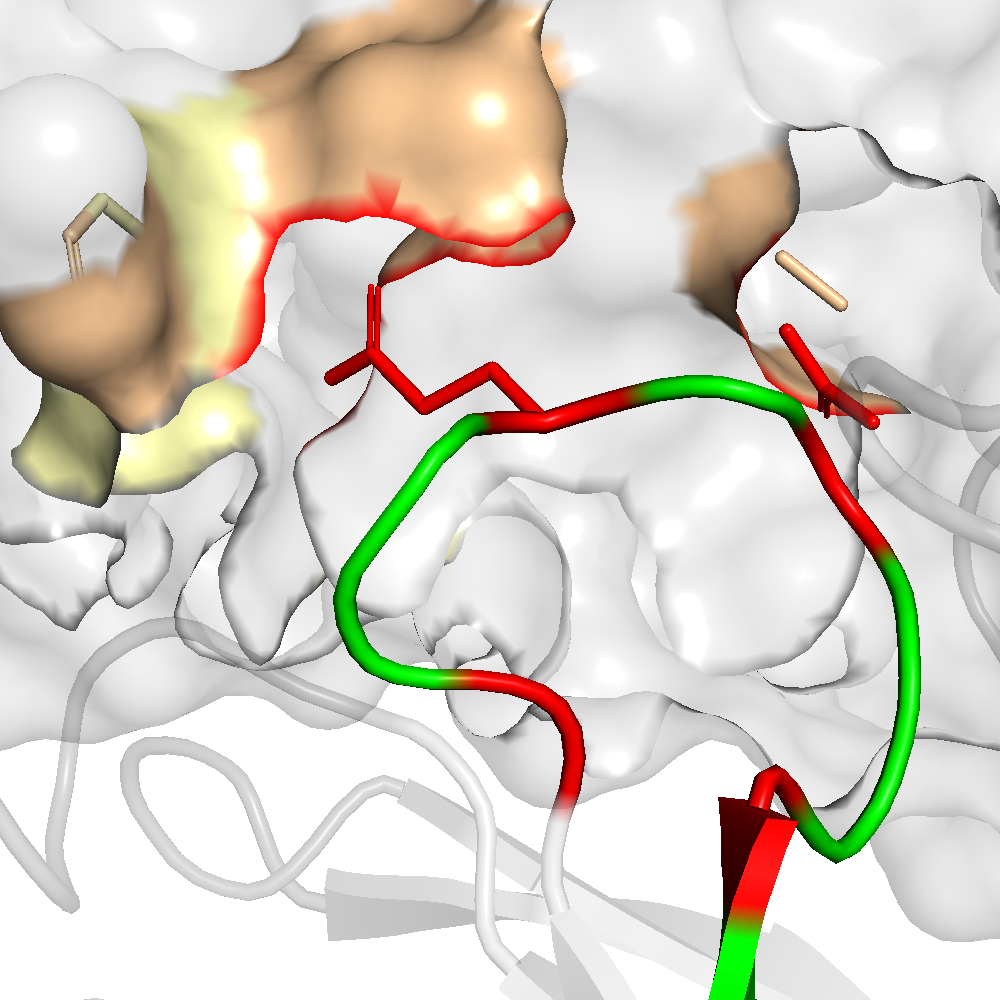}     \captionsetup{justification=centering}
    \caption{Reference H3 \\$E_{total} =-1.62 \text{ kcal/mol}$ \\$\Delta G=58.58 \text{ kcal/mol}$}
  \end{subfigure}
  \hfill
  \begin{subfigure}{0.24\textwidth}
   \includegraphics[width=\linewidth]{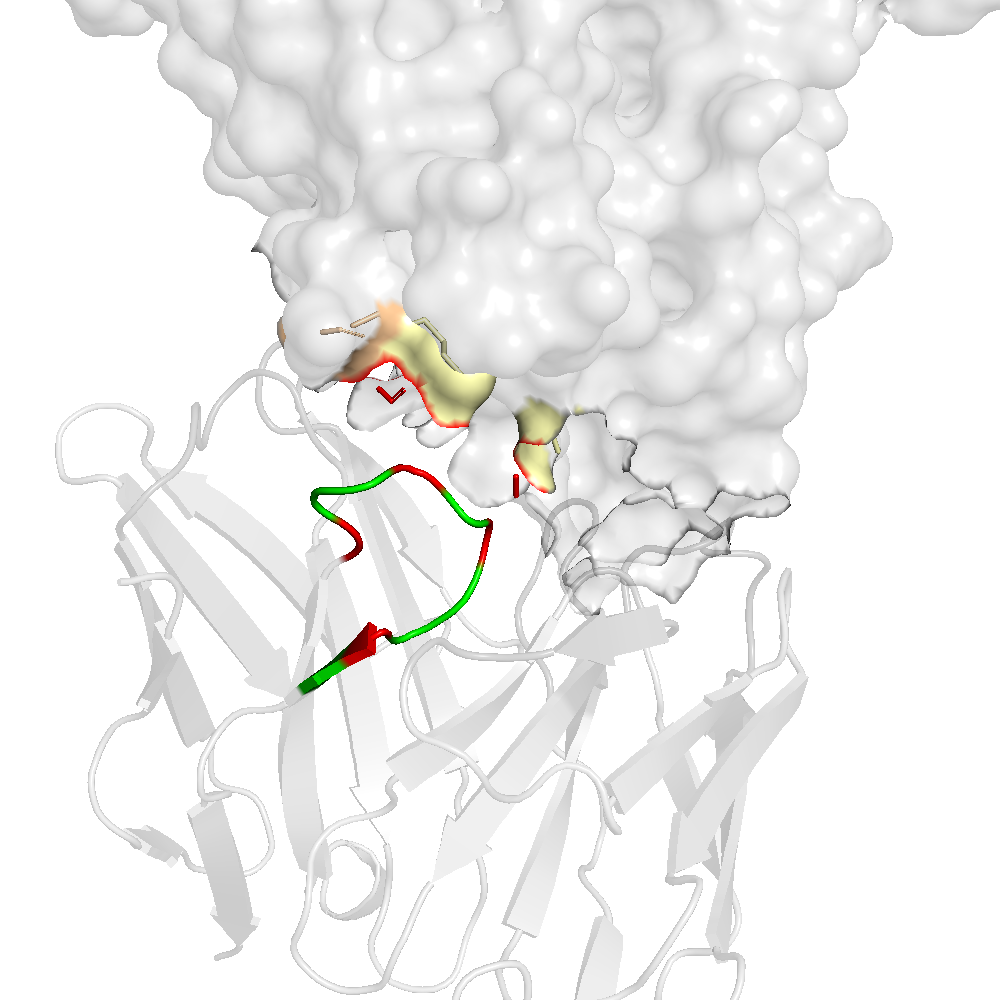}  \captionsetup{justification=centering}
    \caption{Reference \\ Antigen-Antibody\\Complex}
  \end{subfigure}
  \caption{\textit{De novo} Generated and Reference CDR-H3s for \texttt{4dvr} complex.  For DiffAb and AbFlowNet, we generated 100 CDRs and selected the one with the lowest $\Delta \text{G}$.}
  \label{fig:4dvr_main}
\vspace{-1em}
\end{figure*}

Figure \ref{fig:4dvr_main} visualizes \href{https://www.rcsb.org/structure/4DVR}{PDB entry 4DVR} (HIV-1 gp120 CD4-binding site \citep{4dvr_paper}). Hydrophobic residues are colored orange (antigen) and red (CDR); hydrophilic residues are yellow and green. Specificity favors orange-red and yellow-green pairings. AbFlowNet's CDR-H3 (b) is embedded deeper in the antigen than even the reference (c) and has fewer regions of poor specificity than DiffAb (a), which generates a nearly all-hydrophobic CDR. Additional qualitative examples are in Appendix \ref{app:qual}.

\section{Discussion}
\label{sec:discussion}
\vspace{-0.5em}

\paragraph{Comparison with State-of-the-Art Architectures}
AbFlowNet contributes a training objective and leaves the diffusion backbone architecture untouched. State-of-the-art CDR design models -- AbX \citep{abx}, AbNovo \citep{abnovo}, IgGM \citep{iggm} -- contribute architectural improvements instead, typically attaching a billion-scale Protein Language Model \citep{esm2} to a $\sim$$3\times$ larger diffusion trunk; they achieve higher absolute AAR and RMSD than diffusion-only models such as DiffAb, AbFlowNet, and AbDPO (Appendix Table~\ref{tab:sota}). Our GFlowNet-diffusion framework applies to any backbone whose denoising trajectory is differentiable, including architectures that incorporate large frozen pretrained PLMs. We validate AbFlowNet's methodology on DiffAb due to training-code availability, demonstrating consistent improvements over the baseline and concrete advantages over post-training RL frameworks such as AbDPO. AbFlowNet can be readily integrated into PLM-based diffusion architectures: trajectory rollouts on training data are independently parallelizable across GPUs, and backpropagation runs over a single random timestep per update, so peak memory is comparable to standard diffusion training of the same backbone.

\paragraph{Toward higher-fidelity rewards.} Because AbFlowNet only requires energies for the fixed training set, it sidesteps the dominant cost of online-RL methods like AbDPO, which score $\sim$556K synthetic CDRs over training (vs.\ AbFlowNet's 9{,}410, a $\sim$59$\times$ reduction; Appendix \ref{app:runtime}). The same property makes it tractable to upgrade the reward beyond Rosetta \texttt{InterfaceAnalyzer} \citep{chaudhury2010pyrosetta} -- to high-fidelity simulators such as \texttt{OpenMM Yank} \citep{yank2020} that take hours per complex, or to \textit{in vitro} affinity data \citep{binding_energy_iti, binding_energy_spr1, binding_energy_ocetet} for the $\sim$736 SAbDab complexes that have it. We use Rosetta in this paper for parity with prior work, and confirm that gains transfer to the orthogonal FoldX \citep{foldx, desautels_foldx} in Sec. \ref{ssec:orthogonal}.

\section{Conclusion}
\vspace{-0.5em}
We presented AbFlowNet, a novel framework integrating Diffusion Models and GFlowNets for antibody CDR design. AbFlowNet directly incorporates binding energy signals throughout training, jointly optimizing sequence/structure generation and binding affinity. This approach avoids the trade-offs seen in RL-based methods, such as strong reliance on \textit{in silico} binding energy estimation and usage of test set data. AbFlowNet outperforms its base diffusion model (DiffAb) in all metrics and is competitive with expensive RL approaches while only using precomputed rewards.

\section{Acknowledgements}
This work was supported in part by the Google CAHSI grant and the National Science Foundation (NSF) under Grant No. 2431569.

\bibliography{main}
\bibliographystyle{unsrtnat}

\newpage
\appendix
\begin{center}
  \Large Technical Appendices and Supplementary Material
\end{center}

\section{Performance on Reference CDR Reconstruction}
\label{rabd_results}

\begin{table}[ht]
\centering
\caption{Evaluation of the generated antibody CDRs (sequence-structure co-design) on the RAbD test dataset (60 sequences) using AAR (\%), RMSD (\AA), and IMP (\%) metrics.}
\vspace{-1em}
\resizebox{1.00\textwidth}{!}{%
\begin{minipage}{0.57\linewidth}
\centering
\subcaption*{}
\begin{tabular}{@{}llccc@{}}
\toprule
CDR & Method & AAR$\uparrow$ & RMSD$\downarrow$  & IMP$\uparrow$  \\
\midrule
H1 & Diffab    & \textbf{64.23\%} & 1.153\AA & 69.27\% \\
   & AbFlowNet & 63.49\% & \textbf{0.974\AA} & \textbf{73.66\%} \\
\midrule
H2 & Diffab    & 35.87\% & 1.095\AA & 46.79\% \\
   & AbFlowNet & \textbf{38.06\%} & \textbf{0.848\AA} & \textbf{60.07\%} \\
\midrule
H3 & Diffab    & 24.34\% & 3.236\AA & \textbf{14.38\%} \\
   & AbFlowNet & \textbf{25.08\%} & \textbf{3.194\AA} & 12.65\% \\
\bottomrule
\end{tabular}
\end{minipage}%

\begin{minipage}{0.57\linewidth}
\centering
\subcaption*{}
\begin{tabular}{@{}llccc@{}}
\toprule
CDR & Method & AAR $\uparrow$ & RMSD $\downarrow$  & IMP $\uparrow$  \\
\midrule
L1 & DiffAb    & 53.69\% & 1.153\AA & 55.51\% \\
   & AbFlowNet & \textbf{55.62\%} & \textbf{0.974\AA} & \textbf{56.83\%} \\
\midrule
L2 & DiffAb    & 50.46\% & 0.795\AA & 68.78\% \\
   & AbFlowNet & \textbf{54.09\%} & \textbf{0.782\AA} & \textbf{70.64\%} \\
\midrule
L3 & DiffAb    & \textbf{44.87\%} & 3.840\AA & \textbf{36.98\%} \\
   & AbFlowNet & 44.68\% & \textbf{1.310\AA} & 34.70\% \\
\bottomrule
\end{tabular}
\label{tab:rabd_results}
\end{minipage}
}
\end{table}

Table \ref{tab:rabd_results} shows the performance of AbFlowNet and the baseline diffusion model DiffAb on the RaBD dataset. Both models were trained with identical hyperparameters and the same number of gradient updates; the only difference is that AbFlowNet incorporates the TB objective from Eq. \ref{eqn:loss_combo}.

\section{Experimental Setup Details}

\subsection{Hardware Specifications and Runtime}
\label{app:hardware}
We conduct all experiments using a Linux machine with Intel(R) Xeon(R) Silver 4314 CPU with 512GB memory and one NVIDIA RTX A6000 48GB GPU. Training the first $195,000$ steps without the GFlowNet TB objective took $\sim27$ hours, and training the last $5,000$ steps took $\sim18$ hours. Sampling 100 times for each CDR region for every complex in the RAbD test dataset took $\sim12$ hours.

\subsection{Balancing Between Diffusion and Trajectory Balance Objectives}
\label{app:ablate_w}

\begin{figure}[htbp]
    \centering
    \begin{subfigure}[b]{0.7\textwidth}
        \centering
        \includegraphics[width=\linewidth]{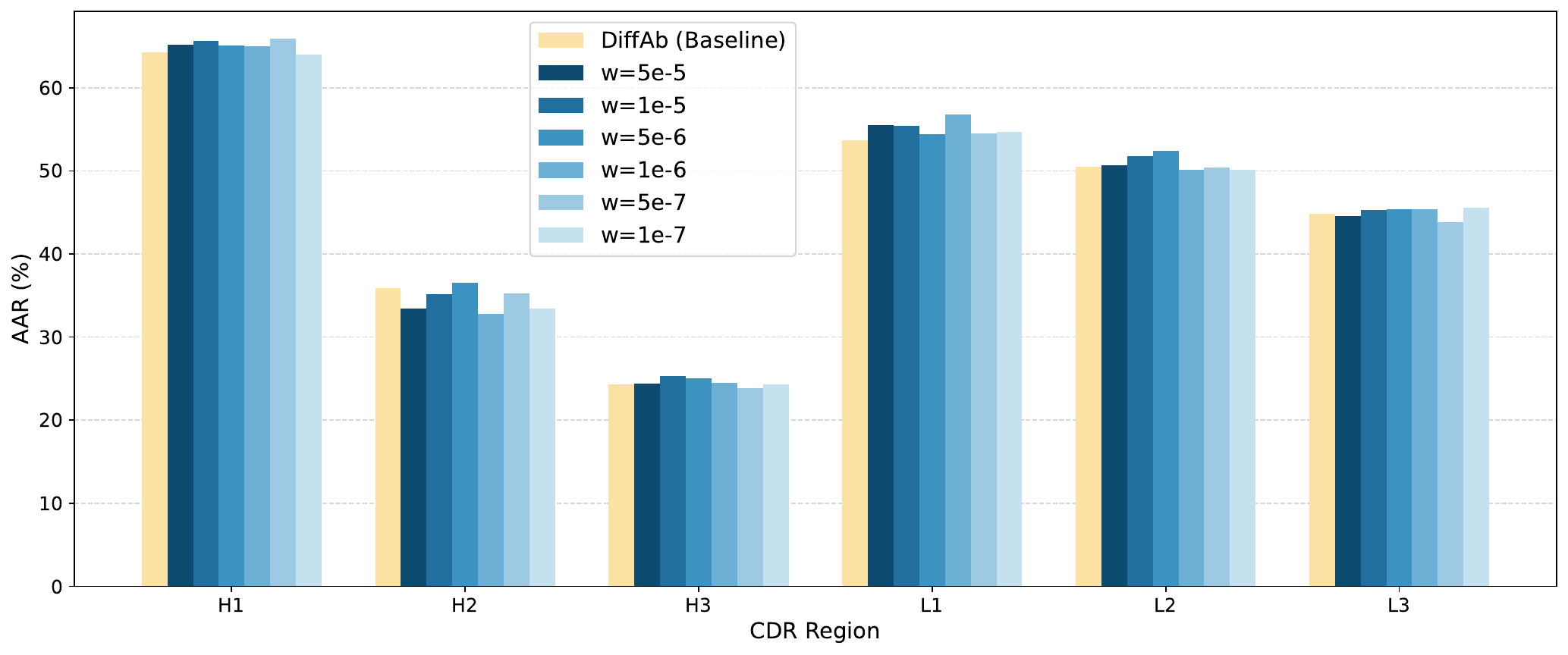}
        \caption{Amino Acid Recovery (ARR) Rate Comparison. Higher is better.}
    \end{subfigure}

    \vspace{0.0cm}

    \begin{subfigure}[b]{0.7\textwidth}
        \centering
        \includegraphics[width=\linewidth]{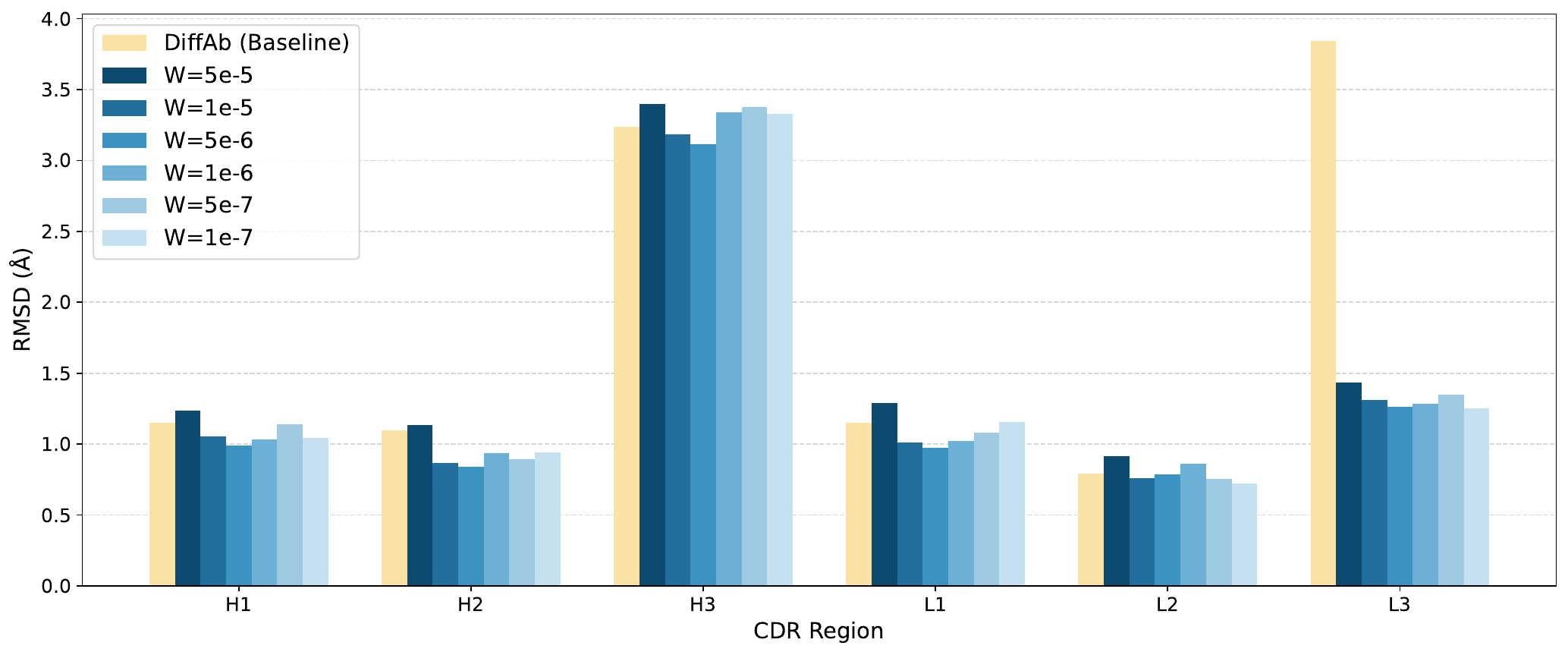}
        \caption{Root Mean Square Deviation (RMSD) Comparison. Lower is better.}
    \end{subfigure}

    \caption{Hyperparameter search for TB loss weight $w$ in Eqn. \ref{eqn:loss_combo} on the RAbD \citep{rabd} dataset. The RMSD of DiffAb on the L3 CDR region is significantly worse than that of AbFlowNet. We retrained and evaluated DiffAb and AbFlowNet using a different seed to confirm this discrepancy (retrained DiffAb RMSD $4.06$ \AA and retrained AbFlowNet RSMD $1.362$).}
    \label{fig:loss_weight}
\end{figure}

Although the method for computing the forward and backward flow of rewards is computationally expensive, the final trajectory balance loss is simply added to the diffusion reconstruction losses, as shown in Eqn. \ref{eqn:loss_combo}. The Trajectory Balance (TB) loss is typically ranges from $10^4$ to $10^6$, while the three diffusion losses have magnitudes between 0 and 1 after $195000$ training steps. This necessitates a TB loss weight $w$ to balance between the flow matching and reconstruction objectives. We train and test AbFlowNet with $w$ ranging from $5e-5$ to $1e-7$ and find that learning rates between $1e-5$ and $1e-6$ are consistently better than the baseline set by DiffAb. Detailed results are shown in Figure \ref{fig:loss_weight}.

\subsection{Post-Training with Trajectory Balance}
\label{app:ablate_time_step}

So far, we have focused on matching the number of gradient updates between the DiffAb baseline and AbFlowNet to isolate the effect of optimizing binding energy via trajectory balance.

\begin{figure}[htbp]
    \centering
    \begin{subfigure}[b]{0.7\textwidth}
        \centering
        \includegraphics[width=\linewidth]{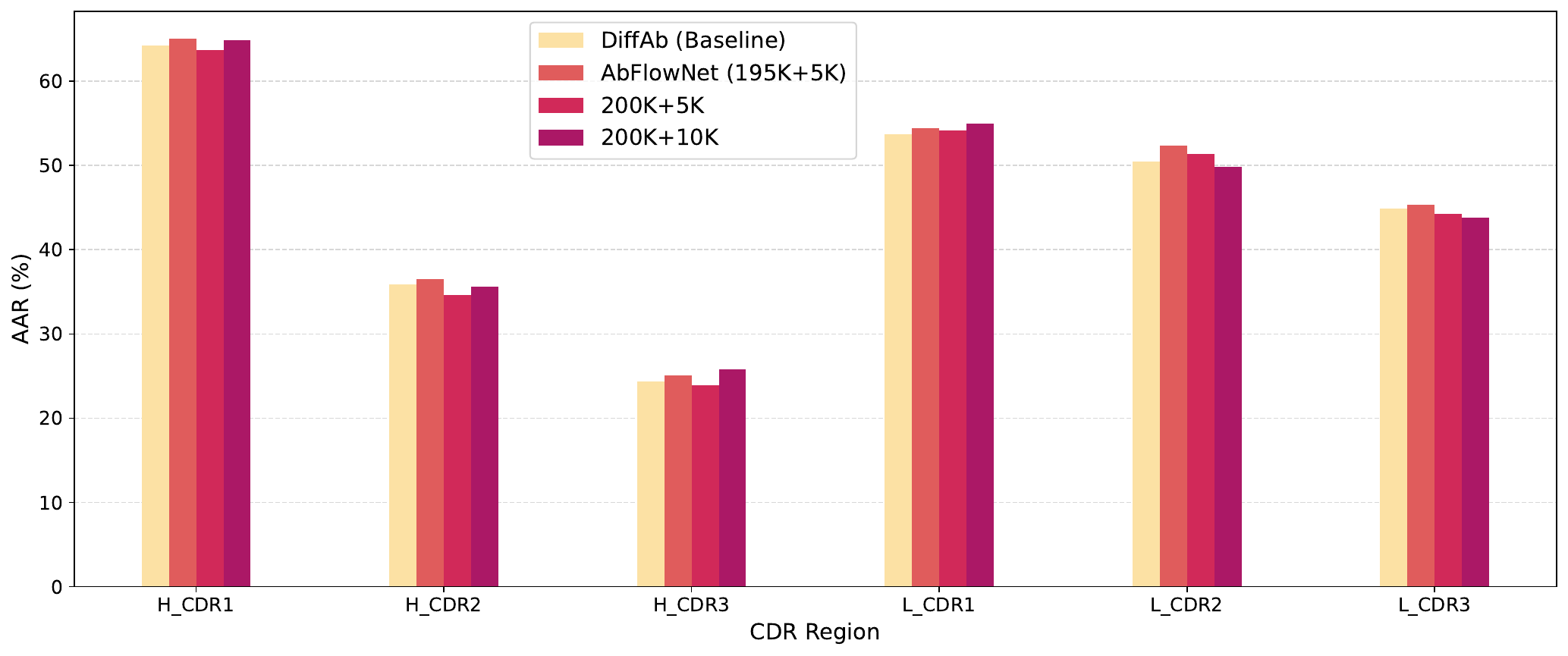}
        \caption{Amino Acid Recovery (ARR) Rate Comparison. Higher is better.}
    \end{subfigure}

    \vspace{0.0cm}

    \begin{subfigure}[b]{0.7\textwidth}
        \centering
        \includegraphics[width=\linewidth]{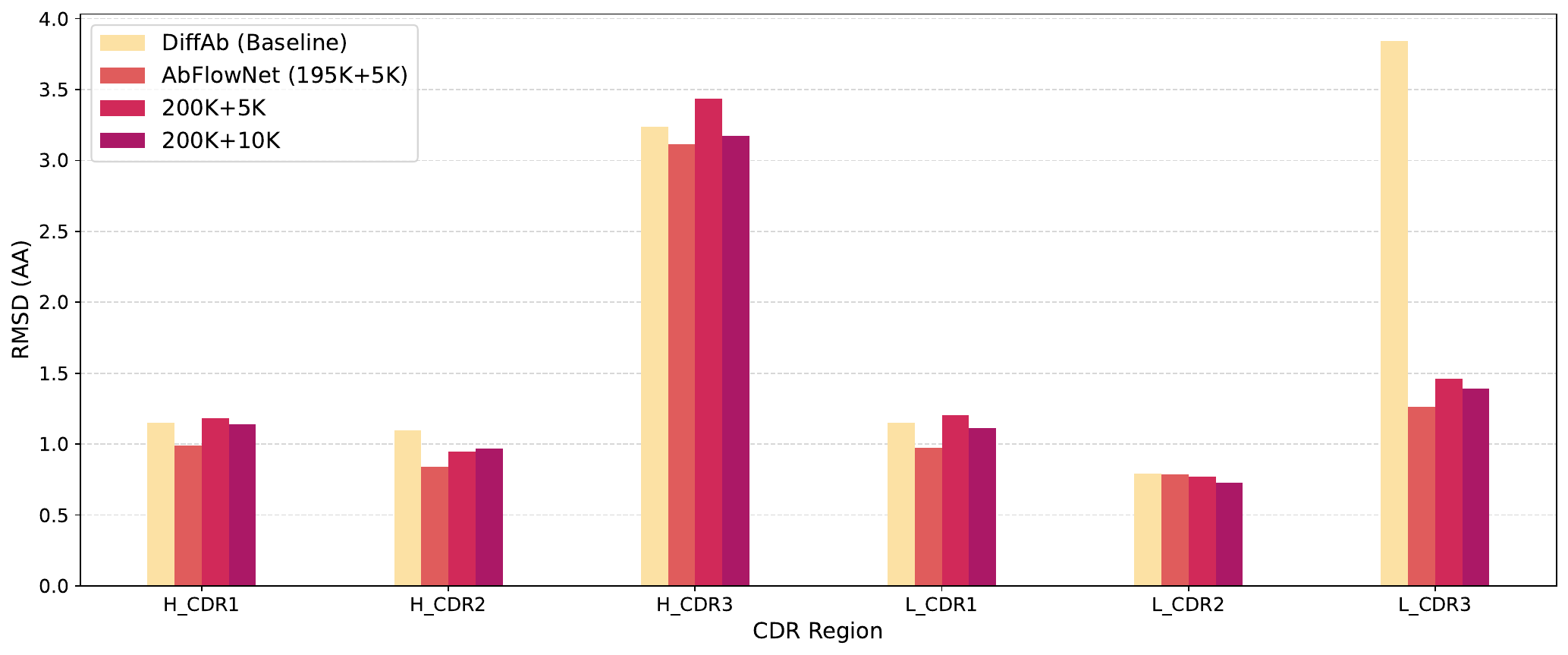}
        \caption{Root Mean Square Deviation (RMSD) Comparison. Lower is better.}
    \end{subfigure}

    \caption{Training Diffusion+GFlowNet models with different training steps on the RAbD dataset \citep{rabd}. Separating the reconstruction and flow matching steps do not meaningfully improve performance over AbFlowNet.}
    \label{fig:steps_ablation}
\end{figure}

However, training with sparse feedback is often framed as a separate stage after unsupervised learning \citep{abdpo, db_on_images}. We test this setup by first training the diffusion model on only the reconstruction objectives for 200K steps and a further 10K steps with the weighted TB loss enabled ($w=5e-6$). We find that training beyond 200K steps with reconstruction objectives generally tends to overfit the dataset, while training with only the TB loss objective harms metrics such as AAR and RMSD, similar to the findings of \citet{abdpo}. Detailed results are shown in Figure \ref{fig:steps_ablation}.

\clearpage

\section{Comparison with State-of-the-art CDR Design Architectures}

\begin{table}[ht]
\centering
\caption{Evaluation of state-of-the-art CDR design methods on the RAbD test dataset. The DiffAb baseline has been independently benchmarked using different parameters by AbDPO \citep{abdpo}, AbNovo \citep{abnovo}, and IgGM \citep{iggm}, which we denote in parentheses.}
\vspace{-1em}
\resizebox{1.00\textwidth}{!}{%
\begin{minipage}{0.5\linewidth}
\centering
\subcaption*{}
\begin{tabular}{@{}llcc@{}}
\toprule
CDR & Method & AAR$\uparrow$ & RMSD$\downarrow$ \\
\midrule
H1 & DiffAb (Ours) & 64.23 & 1.153 \\
   & AbFlowNet & 63.49 & 0.974 \\
   \cmidrule(lr){2-4}
   & DiffAb (AbDPO) & - & - \\
   & AbDPO & - & - \\
   \cmidrule(lr){2-4}
   & DiffAb (AbNovo) & 70.01 & 0.88 \\
   & AbX & 80.92 & 0.85 \\
   & AbNovo & 85.25 & 0.66 \\
   \cmidrule(lr){2-4}
   & DiffAb (IgGM) & 63.7 & 0.623 \\
   & IgGM & 74.0 & 0.555 \\
\midrule
H2 & DiffAb (Ours) & 35.87 & 1.095 \\
   & AbFlowNet & 38.06 & 0.848 \\
   \cmidrule(lr){2-4}
   & DiffAb (AbDPO) & - & - \\
   & AbDPO & - & - \\
   \cmidrule(lr){2-4}
   & DiffAb (AbNovo) & 38.52 & 0.78 \\
   & AbX & 70.73 & 0.76 \\
   & AbNovo & 78.56 & 0.76 \\
   \cmidrule(lr){2-4}
   & DiffAb (IgGM) & 39.4 & 0.586 \\
   & IgGM & 64.4 & 0.486 \\
\midrule
H3 & DiffAb (Ours) & 24.34 & 3.236 \\
   & AbFlowNet & 25.08 & 3.194 \\
   \cmidrule(lr){2-4}
   & DiffAb (AbDPO) & 36.42 & 2.34 \\
   & AbDPO & 26.48 & 2.48 \\
   \cmidrule(lr){2-4}
   & DiffAb (AbNovo) & 28.05 & 2.86 \\
   & AbX & 44.18 & 2.50 \\
   & AbNovo & 49.93 & 2.19 \\
   \cmidrule(lr){2-4}
   & DiffAb (IgGM) & 22.6 & 2.646 \\
   & IgGM & 36.0 & 2.131 \\
\bottomrule
\end{tabular}
\end{minipage}%
\begin{minipage}{0.5\linewidth}
\centering
\subcaption*{}
\begin{tabular}{@{}llcc@{}}
\toprule
CDR & Method & AAR$\uparrow$ & RMSD$\downarrow$ \\
\midrule
L1 & DiffAb (Ours) & 53.69 & 1.153 \\
   & AbFlowNet & 55.62 & 0.974 \\
   \cmidrule(lr){2-4}
   & DiffAb (AbDPO) & - & - \\
   & AbDPO & - & - \\
   \cmidrule(lr){2-4}
   & DiffAb (AbNovo) & 61.07 & 0.85 \\
   & AbX & 80.37 & 0.80 \\
   & AbNovo & 84.34 & 0.66 \\
   \cmidrule(lr){2-4}
   & DiffAb (IgGM) & 60.8 & 0.749 \\
   & IgGM & 75.0 & 0.589 \\
\midrule
L2 & DiffAb (Ours) & 50.46 & 0.795 \\
   & AbFlowNet & 54.09 & 0.782 \\
   \cmidrule(lr){2-4}
   & DiffAb (AbDPO) & - & - \\
   & AbDPO & - & - \\
   \cmidrule(lr){2-4}
   & DiffAb (AbNovo) & 58.58 & 0.55 \\
   & AbX & 84.53 & 0.45 \\
   & AbNovo & 88.25 & 0.32 \\
   \cmidrule(lr){2-4}
   & DiffAb (IgGM) & 59.9 & 0.466 \\
   & IgGM & 74.3 & 0.378 \\
\midrule
L3 & DiffAb (Ours) & 44.86 & 3.840 \\
   & AbFlowNet & 44.68 & 1.310 \\
   \cmidrule(lr){2-4}
   & DiffAb (AbDPO) & - & - \\
   & AbDPO & - & - \\
   \cmidrule(lr){2-4}
   & DiffAb (AbNovo) & 47.57 & 1.39 \\
   & AbX & 65.89 & 1.21 \\
   & AbNovo & 73.88 & 0.86 \\
   \cmidrule(lr){2-4}
   & DiffAb (IgGM) & 42.4 & 1.017 \\
   & IgGM & 63.5 & 0.847 \\
\bottomrule
\end{tabular}
\end{minipage}
}
\label{tab:sota}
\end{table}

\section{AbFlowNet Outperforms DiffAb at Every Sampling Budget}
\label{app:sampling_sweep}

The Top-1 metrics in Table \ref{tab:energy_h3} select the best-scoring CDR among $N$ generated samples and are therefore sensitive to the sampling budget. To quantify this, we sweep $N$ from 5 to 5{,}000 on the RAbD test set and compare AbFlowNet to DiffAb on Top-1 CDR-Ag $\Delta G$ (Figure~\ref{fig:sampling_sweep}).

\begin{table}[h]
\centering
\caption{Top-1 CDR-Ag $\Delta G$ (kcal/mol) on the RAbD test set as a function of sampling budget $N$. AbFlowNet remains uniformly better than DiffAb across all budgets.}
\label{tab:sampling_sweep}
\begin{tabular}{r|ccc}
\toprule
$N$ & \textbf{DiffAb} & \textbf{AbFlowNet} & \textbf{Improvement} \\
\midrule
5    & 81.16 & 45.95 & 35.21 \\
10   & 39.38 & 25.25 & 14.13 \\
50   & 18.71 & 3.37  & 15.34 \\
100  & 11.20 & 1.71  & 9.49 \\
500  & 6.51  & -0.10 & 6.61 \\
1000 & 5.59  & -1.15 & 6.74 \\
2528 & 4.02  & -2.94 & 6.96 \\
5000 & 3.11  & -3.64 & 6.75 \\
\bottomrule
\end{tabular}
\end{table}

At the AbDPO budget of $N=2{,}528$, AbFlowNet reaches $\Delta G = -2.94$ kcal/mol, compared to AbDPO's reported $-4.85$. AbDPO closes that gap by sampling 10{,}112 CDRs \emph{per test complex} during preference-pair construction \citep{abdpo}; AbFlowNet trains exclusively on training-set complexes and therefore has no access to test-set energy signals. AbFlowNet's improvement over DiffAb is consistent at every budget, indicating that the gains are not an artifact of any particular $N$.

\section{AbFlowNet's Gains Transfer Through Protein-MPNN Inverse Folding}
\label{app:mpnn}

To disentangle the value of the co-designed sequence from the structure, we replace each generated sequence with the output of \texttt{Protein-MPNN} \citep{proteinmpnn} conditioned on the generated backbone, followed by a \texttt{FastRelax} step to resolve side-chain clashes. We then re-score with FoldX.

\begin{table}[h]
\centering
\caption{FoldX $\Delta G$ (kcal/mol) on the RAbD test set under co-design and structure $\rightarrow$ sequence variants. \textbf{Bold} indicates best per row.}
\label{tab:mpnn}
\resizebox{0.95\textwidth}{!}{%
\begin{tabular}{l|cccc}
\toprule
\textbf{CDR} & \textbf{DiffAb} & \textbf{AbFlowNet} & \textbf{DiffAb-MPNN} & \textbf{AbFlowNet-MPNN} \\
\midrule
H\_CDR1 & -8.00 & \textbf{-9.07} & -7.67 & -8.81 \\
H\_CDR2 & -8.40 & \textbf{-9.60} & -7.68 & -8.86 \\
H\_CDR3 & -2.11 & -3.25         & -2.39 & \textbf{-3.84} \\
L\_CDR1 & -8.46 & -9.30         & -8.43 & \textbf{-9.31} \\
L\_CDR2 & -8.50 & \textbf{-9.54} & -7.82 & -8.83 \\
L\_CDR3 & -7.06 & \textbf{-7.93} & -6.41 & -7.50 \\
\bottomrule
\end{tabular}
}
\end{table}

Two observations follow. First, AbFlowNet beats DiffAb in both the co-design and the structure$\rightarrow$sequence regimes, indicating that the energy gains transfer regardless of how the sequence is produced. Second, co-design wins on H1, H2, L2, and L3, while structure$\rightarrow$sequence wins on the most variable region, CDR-H3 (and marginally on L1). Protein-MPNN is trained on $\sim$200K PDB structures, so it can leverage substantially more sequence diversity than a co-design model trained on $\sim$9K antibody complexes from SAbDab; this is consistent with the H3-specific strength of inverse folding noted in concurrent work \citep{abmpnn}. Germinal \citep{germinal} adopts a similar hybrid intuition (co-design followed by AbMPNN refinement), supporting the view that co-design remains valuable as an auxiliary structural objective even when the final sequence comes from inverse folding.

\section{Train--Test Overlap; AbFlowNet Generalizes to Novel Antigens}
\label{app:novel_antigen}

To rule out memorization, we computed the minimum $C_\alpha$-RMSD and maximum TM-score between each of the 60 RAbD test complexes and all 9{,}410 training complexes. Minimum RMSD is centered around 25\AA{} (range 15--30\AA{}) and the maximum TM-score is below 0.2 -- well under the 0.5 threshold for significant structural similarity. The evaluation regime is therefore non-redundant and structurally distribution-shifted with respect to training.

We additionally identified the subset of test complexes whose antigens do not occur in the training set (``novel antigen'' subset) and report restricted metrics:

\begin{table}[h]
\centering
\caption{Mean Top-1 metrics on the novel-antigen subset of RAbD (CDR-H3, $N=100$). Lower is better.}
\label{tab:novel_antigen}
\begin{tabular}{l|cc}
\toprule
\textbf{Metric} & \textbf{DiffAb} & \textbf{AbFlowNet} \\
\midrule
Mean CDR $E_{\text{total}}$ (kcal/mol) & 503.39 & \textbf{396.19} \\
Mean CDR-Ag $\Delta G$ (kcal/mol)      & 25.47  & \textbf{10.03}  \\
\bottomrule
\end{tabular}
\end{table}

AbFlowNet's gains hold on antigens that the model has not seen in training, supporting the claim that the training objective generalizes beyond memorized contexts.

\section{AbFlowNet Preserves Generation Diversity, Avoiding Mode Collapse}
\label{app:diversity}

A common failure mode of energy-optimizing methods is mode collapse onto a single high-reward sample. We measure mean pairwise sequence identity among 100 generated CDRs per complex, averaged across the RAbD test set. Lower means more diverse generation.

\begin{table}[h]
\centering
\caption{Mean pairwise sequence identity (\%, lower is more diverse).}
\label{tab:diversity}
\begin{tabular}{l|cc}
\toprule
\textbf{CDR} & \textbf{DiffAb} & \textbf{AbFlowNet} \\
\midrule
H\_CDR1 & 70.24 & 70.99 \\
H\_CDR2 & 44.09 & 44.81 \\
H\_CDR3 & 29.51 & 31.65 \\
L\_CDR1 & 58.71 & 61.89 \\
L\_CDR2 & 60.04 & 62.92 \\
L\_CDR3 & 53.20 & 54.53 \\
\midrule
\textbf{Overall} & 52.63 & 54.46 \\
\bottomrule
\end{tabular}
\end{table}

AbFlowNet's pairwise identity is only $\sim$1.8 percentage points higher than DiffAb's overall, indicating that diversity is well preserved despite reward-based optimization. This is consistent with the theoretical property of GFlowNets: trajectory balance enforces that the probability of generating a CDR is \emph{proportional} to its reward, rather than \emph{maximizing} it as RL/DPO would. Concentrating probability on the single highest-reward mode is itself a violation of the TB objective \citep{gflownet1}.

\section{Runtime Decomposition: AbFlowNet Uses 59$\times$ Fewer Energy Calls than AbDPO}
\label{app:runtime}

The dominant runtime cost in energy-optimized antibody design is binding-energy estimation. \texttt{PyRosetta InterfaceAnalyzer} requires $T_{\text{relax}} + T_{\text{energy}} \approx 10.81$ seconds per CDR; per-step forward and backward passes, $T_f + T_b$, are several orders of magnitude faster.

\paragraph{AbDPO.} For each test complex, AbDPO samples $M$ CDRs with full diffusion rollouts, scores each, and uses the resulting preference pairs to fine-tune the model:
\[
\text{Cost}_{\text{AbDPO}} = O\bigl(N_{\text{test}} \cdot \bigl(M \cdot (\text{diffusion\_steps} \cdot T_f + T_{\text{relax}} + T_{\text{energy}}) + \text{iters} \cdot (T_f + T_b)\bigr)\bigr).
\]
With reported values $N_{\text{test}}=55$ and $M=10{,}112$, energy estimation alone requires $55 \cdot 10{,}112 \cdot 10.81\,\text{s} \approx 69.65$ days of single-machine compute (parallelizable, but the total compute is preserved).

\paragraph{AbFlowNet.} Energy is precomputed once for the 9{,}410 training-set complexes and amortized across training:
\[
\text{Cost}_{\text{AbFlowNet}} = O(N_{\text{train}} \cdot T_{\text{energy}}) + O(\text{iters} \cdot (\text{diffusion\_steps} \cdot T_f + T_b)).
\]
Concretely, AbFlowNet trained for 200K total steps (195K diffusion pretraining + 5K GFlowNet fusion) on a single NVIDIA RTX A6000 in $\sim$45 hours, plus a one-time $\sim$28 hours of energy precomputation across the training complexes. The number of energy-estimation calls is reduced by $\sim$59$\times$ (9{,}410 vs. $\sim$556K) relative to AbDPO, and -- crucially -- those calls run on \emph{training} complexes, not test complexes, eliminating one source of test-set leakage. The per-sample inference cost is identical to DiffAb.

\section{DiffAb Test Set Performance}
\label{app:sabdab}

The authors of DiffAb \citep{diffab} proposed a test set consisting of 19 complexes with antigens from several well-known pathogens, including SARS-CoV-2, MERS, influenza, and so on. Since these complexes are part of the SAbDab dataset \citep{dunbar2014sabdab} used for training, we filter our training complexes against the test set using a CDR-H3 sequence identity threshold of 50\%. We retrain both DiffAb and AbFlowNet with this new filtered training set.

\begin{table}[ht]
\centering
\caption{Evaluation of the generated antibody CDRs (sequence-structure co-design) on the DiffAb test dataset (19 sequences).}
\vspace{-2em}
\resizebox{1.00\textwidth}{!}{%
\begin{minipage}{0.55\linewidth}
\centering
\subcaption*{}
\begin{tabular}{@{}llccc@{}}
\toprule
CDR & Method & AAR$\uparrow$ & RMSD$\downarrow$  & IMP$\uparrow$  \\
\midrule
H1 & DiffAb & 68.29\% & 1.090\AA & 53.96\% \\
   & AbFlowNet & \textbf{69.41\%} & \textbf{0.944\AA} & \textbf{63.50\%} \\
\midrule
H2  & DiffAb & 35.94\% & \textbf{0.804\AA} & \textbf{55.60\%} \\
   & AbFlowNet & \textbf{36.39\%} & 0.862\AA & 54.38\% \\
\midrule
H3 & DiffAb & 26.53\% & \textbf{3.183\AA} & \textbf{12.75\%} \\
   & AbFlowNet & \textbf{26.66\%} & 3.321\AA & 8.75\% \\
\bottomrule
\end{tabular}
\end{minipage}%
\hfill
\begin{minipage}{0.55\linewidth}
\centering
\subcaption*{}
\begin{tabular}{@{}llccc@{}}
\toprule
CDR & Method & AAR$\uparrow$ & RMSD$\downarrow$  & IMP$\uparrow$  \\
\midrule
L1  & DiffAb  & 54.40\% & \textbf{0.960\AA} & 62.48\% \\
   & AbFlowNet & \textbf{55.97\%} & 1.019\AA & \textbf{64.84\%} \\
\midrule
L2  & DiffAb  & 42.55\% & \textbf{0.735}\AA & 80.61\% \\
   & AbFlowNet & \textbf{45.52\%} & 0.757\AA & \textbf{84.47\%} \\
\midrule
L3 & DiffAb & 46.15\% & 1.127\AA & 37.07\% \\
   & AbFlowNet & \textbf{46.19\%} & \textbf{1.180\AA} & \textbf{37.26\%} \\
\bottomrule
\end{tabular}
\label{tab:diffab_test_results}
\end{minipage}
}
\end{table}

\begin{table}[h]
\centering
\caption{Summary of Top-1 CDR $E_{total}$ and CDR-Ag $\Delta G$ (kcal/mol) of CDR-H3's designed by DiffAb and AbFlowNet on the DiffAb Test Dataset. ( $\downarrow$ ) denotes that a smaller number is better.}
\begin{tabular}{l|c|rcrcc}
\toprule
\textbf{Methods} & \textbf{\# Samples} & \textbf{CDR} & \textbf{$\mathbf{+E_{total}}$} &\textbf{CDR-Ag} & \textbf{$\mathbf{+\Delta G}$} &\textbf{Test Set}\\
 & & $\mathbf{E_{total}}$ ( $\downarrow$ ) & \textbf{(\%)} & $\mathbf{\Delta G}$ ( $\downarrow$ ) & \textbf{(\%)} & \textbf{Used}\\
\midrule
Reference      && 1.63 && -4.80 & \\
\midrule
DiffAb        & \multirow{2}{*}{100} & 26.33   &     & 11.50&\\
AbFlowNet (Ours)     && 4.23&  89.5  & 1.47 & 149.7 & No\\
\bottomrule
\end{tabular}
\label{tab:diffab_energy_h3}
\end{table}

\section{Approaches to Determining Side-Chain Orientation with Neural Networks}
\label{app:side_chain_ori}

Most generative methods—including DiffAb, AbDPO, AbFlowNet, AlignAb \citep{alignab}, AbX \citep{abx}, AbNovo \citep{abnovo}, etc.—generate only the backbone structure. The orientation of the amino-acid chains isn’t generated through a diffusion process, since it must follow structural constraints such as avoiding overlaps. Instead, we rely on a side-chain packing algorithm such as \texttt{PyRosetta PackRotamerMover} to find the ideal orientation of side-chains. The GNN-based dyMEAN \citep{dymean} jointly generates the backbone and side-chain orientations, but Table \ref{tab:energy_h3} shows that this approach underperforms using \texttt{PyRosetta}. There has been notable research into generating only side-chains conditioned on the backbone with diffusion neural networks \citep{zhang2023diffpack, zhang2024packdock}, and given the limitations, it is a critical direction of future research.

\section{Efficiency Considerations of Diffusion-GFlowNet Fusion}
\label{app:efficiency}
Computing the Trajectory Balance (TB) objective requires complete trajectories, i.e., sampling from the initial Gaussian state to the final denoised CDR conditioned on the antigen-antibody complex. Following \citep{diffab, abdpo, abx}, we use 100 denoising steps. The wall-clock time for 100 denoising steps with a batch size of 16 is $\sim\!20$ seconds, which dominates the run-time of training AbFlowNet. Furthermore, we do backward propagation for only one random time step since storing the activations for all 100 steps is computationally infeasible. This deviates from the original TB \citep{gflownet1} formulation and hence is only an approximation. Following \citep{db_on_images}, we discuss our attempt at using an alternative GFlowNet optimization objective, Detailed Balance, in Appendix \ref{app:DB}.

\section{Detailed Balance Objective}
\label{app:DB}
Detailed Balance (DB) \citep{gflownet1} is an alternative training objective to the standard flow matching constraint and Trajectory Balance, which doesn't require enumerating states or sampling complete trajectories. Rather, DB requires the forward flow from state $s$ to $s'$, $F(s)P_F(s'|s)$ to match the backward flow $F(s')P_B(s|s')$. Concretely, the DB objective is

\begin{equation}
\label{db_objective}
F(s)P_F(s'|s) = F(s')P_B(s_{t-1}|s_t)
\end{equation}

However, the flow of a nonterminal state $s$ is generally not tractable, and hence it is parameterized with a neural network $F_\phi(\cdot)$.
The forward and backward transition probabilities of the entire CDR $S^t$ are $p(S^{t-1}|S^{t}) = \prod_{j=l}^{l+m} p(s^{t}_j|s^{t-1}_j)$ and $q(S^{t}|S^{t-1}) = \prod_{j=l}^{l+m} q(s^{t}_j|s^{t-1}_j)$ respectively. Therefore, the final DB objective is:

\vspace{-1.5em}
\begin{equation}
L^t_{\text{DB}} =\left(\log\frac{F_\phi(S^t)p(S^t|S^{t-1};\theta)}{F_\phi(S^{t-1})q(S^{t-1}|S^t)}\right)^2
\end{equation}

\paragraph{Pilot Attempt Using DB Objective}

In the GFlowNet framework, there are three equivalent optimization objectives: Flow Matching (FM), Detailed Balance (DB), and Trajectory Balance (TB) - each with its own tradeoffs. Flow Matching requires enumerating states and enforcing parity between incoming and outgoing flows. FM is not applicable since the number of states in diffusion models is infinite. We attempt to use a DB that only requires computing the forward and backward flow between two states and enforcing parity.

Similar to optimizing the TB objective outlined in Section \ref{method_tb}, we uniformly sample a timestep $t$ to compute $L^t_{\text{type}}$, $L^t_{\text{pos}}$ and $L^t_{\text{ori}}$. Since we require adjacent state pairs to compute DB, we do a single step of denoising to obtain $S^{t-1}$ from $S^{t}$. To enforce the DB objective \ref{db_objective}, we must compute $F(S^{t})$ and $F(S^{t-1})$. However, intermediate states $S^{t-1}$ and $S^{t}$ are noisy and therefore are not appropriate to be evaluated by a reward function, which would give noisy results. Following \citet{db_on_images}, we define the linearization.

\vspace{-1.5em}
\begin{equation}
\label{eqn:db_approximation}
F_\phi(S^t) = \tilde{F}_\phi(S^t)R(\hat{S^0}) =  \tilde{F}_\phi(S^t)R(FullDenoise_\theta({S^t}))
\end{equation}

where $\tilde{F}_\phi()$ is a scalar function that scales the reward of the estimated fully denoised state. Being a diffusion model, we can fully denoise any noisy state, albeit with a sacrifice in quality.

\paragraph{Key Bottlenecks} At this stage, we run into the key issue that precludes the use of DB in CDR design. We need to compute the energy of the designed CDR using a tool such as \texttt{InterfaceAnalyzer} in the \texttt{Rosetta} \citep{chaudhury2010pyrosetta} software package. \texttt{InterfaceAnalyzer} requires the designed antigen-antibody structure to be complete with side-chains. However, diffusion models generally generate only the backbone and rely on a search-based side-chain packing algorithm such as \texttt{PackRotamerMover}. Both \texttt{PackRotamerMover} and \texttt{InterfaceAnalyzer} are CPU-based utilities, and it takes 10.81 seconds to process a single CDR .pdb file. Determining the energy for the two states for each item in the mini-batch (16 in our experiments) requires $\sim93$ seconds even when parallelized over a 32+ core machine, including multiple data migration costs between the GPU, CPU, and disk. This is in contrast to the millisecond-scale time required for the forward and backward passes.

Therefore, the training runtime is dominated by the time it takes to compute the binding energy reward, and training becomes infeasible.

\paragraph{Neural Surrogate for Rosetta’s InterfaceAnalyzer} To the best of our knowledge, a neural network alternative to \texttt{PackRotamerMover} and \texttt{InterfaceAnalyzer} does not exist. We tried to train a transformer-based neural network to simulate the function of \texttt{InterfaceAnalyzer} directly from the output of the diffusion model. However, this neural network had very low agreement with the \texttt{InterfaceAnalyzer} tool (Pearson's coefficient $0.21$), which itself is unreliable \citep{binding_energy_survey}. This is expected since side-chains play a central role in determining binding affinity \citep{cdr_basics}. Another drawback of the DB objective is the need to compute the binding energy of generated CDRs, which are not guaranteed to be geometrically plausible.

In light of our findings, we finally committed to Trajectory Balance as the only feasible objective for training AbFlowNet despite the need to sample full trajectories.

\section{Evolutionary Plausibility of Generated CDRs: Generated CDRs Cluster with References in ESM-2 Space}
\label{app:evoplaus}

\begin{figure}[ht]
    \centering
    \begin{subfigure}[b]{0.3\textwidth}
        \centering
        \includegraphics[width=\textwidth]{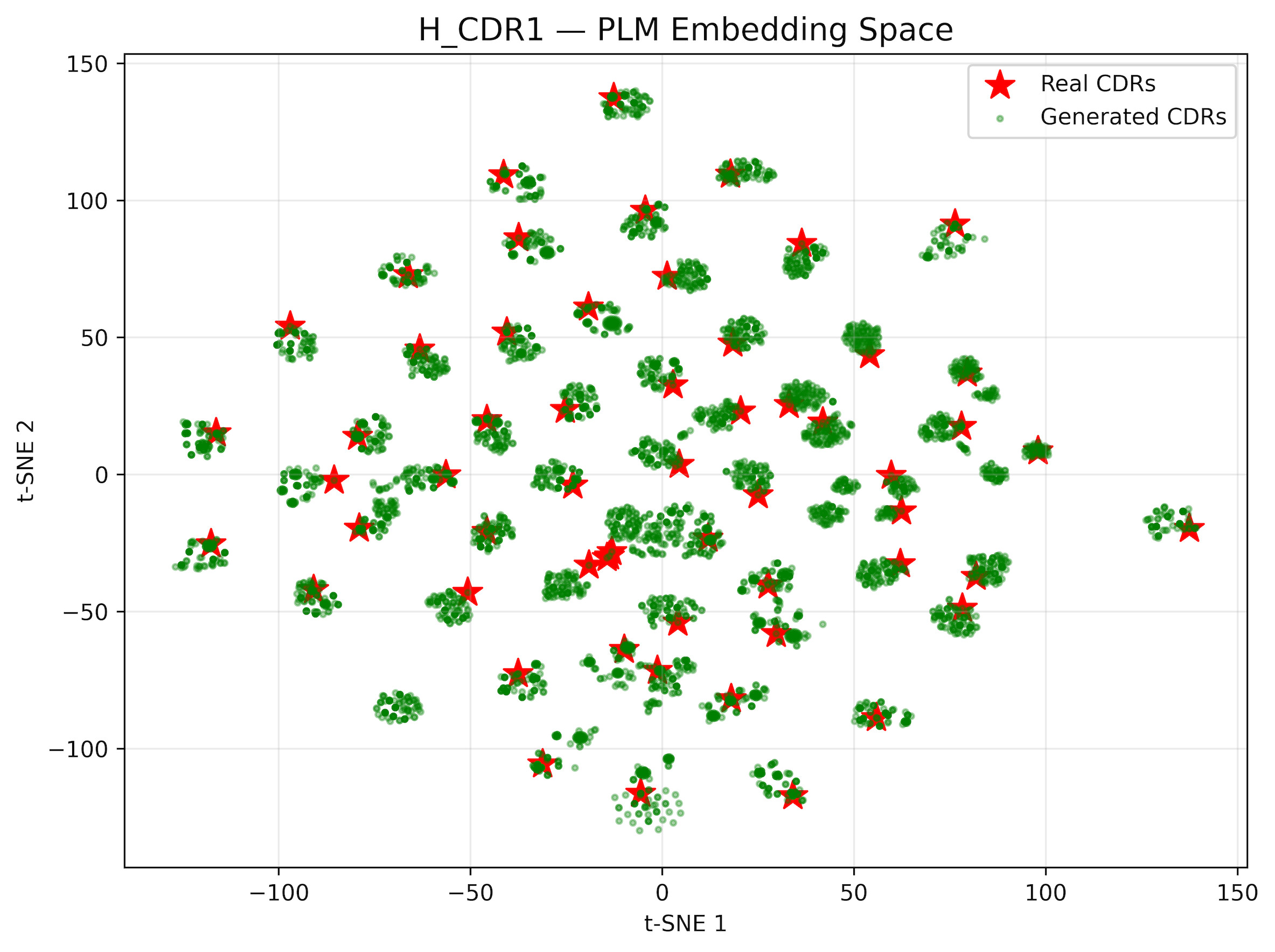}
        \caption{CDR-H1}
    \end{subfigure}
    \hfill
    \begin{subfigure}[b]{0.3\textwidth}
        \centering
        \includegraphics[width=\textwidth]{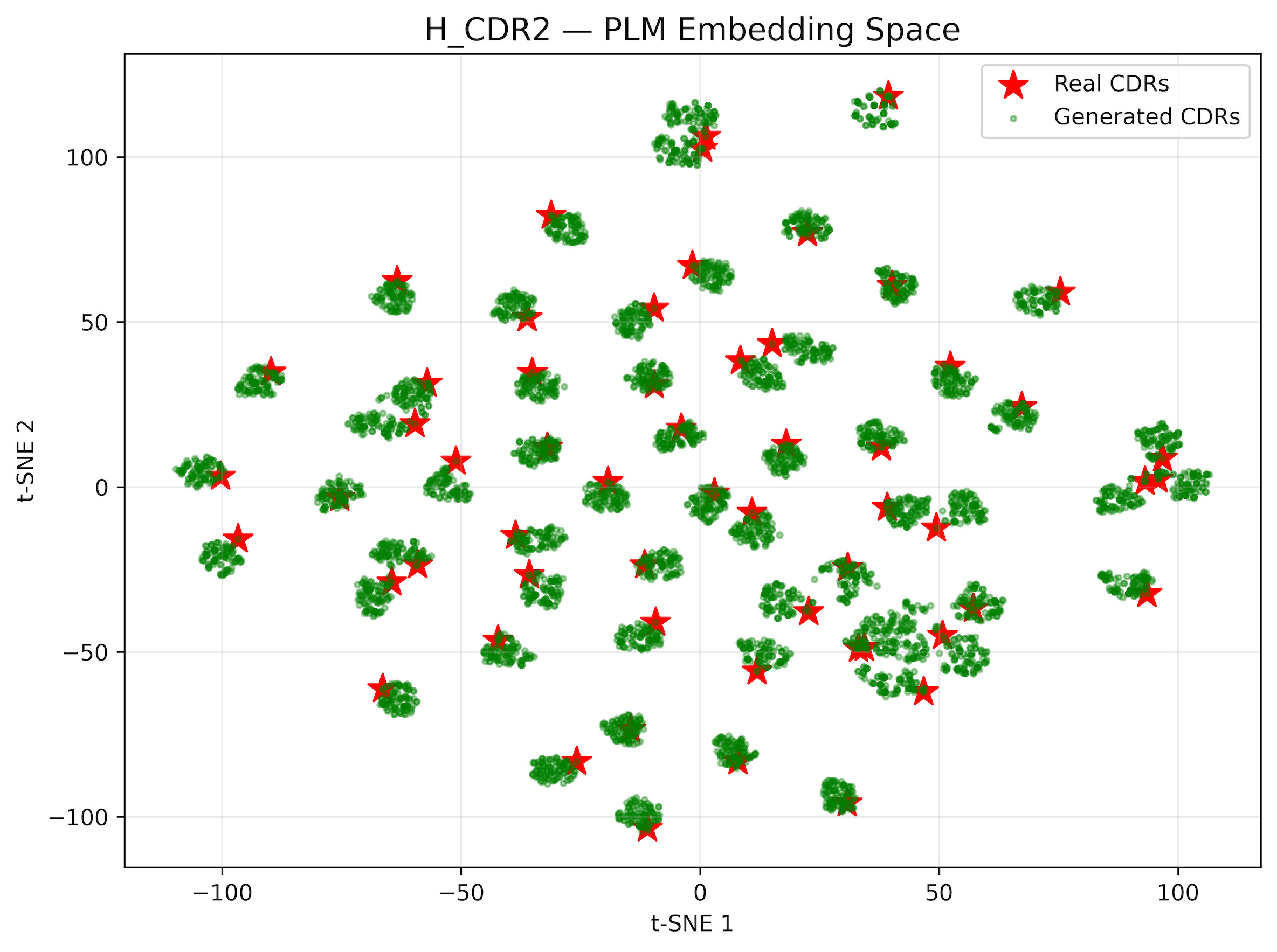}
        \caption{CDR-H2}
    \end{subfigure}
    \hfill
    \begin{subfigure}[b]{0.3\textwidth}
        \centering
        \includegraphics[width=\textwidth]{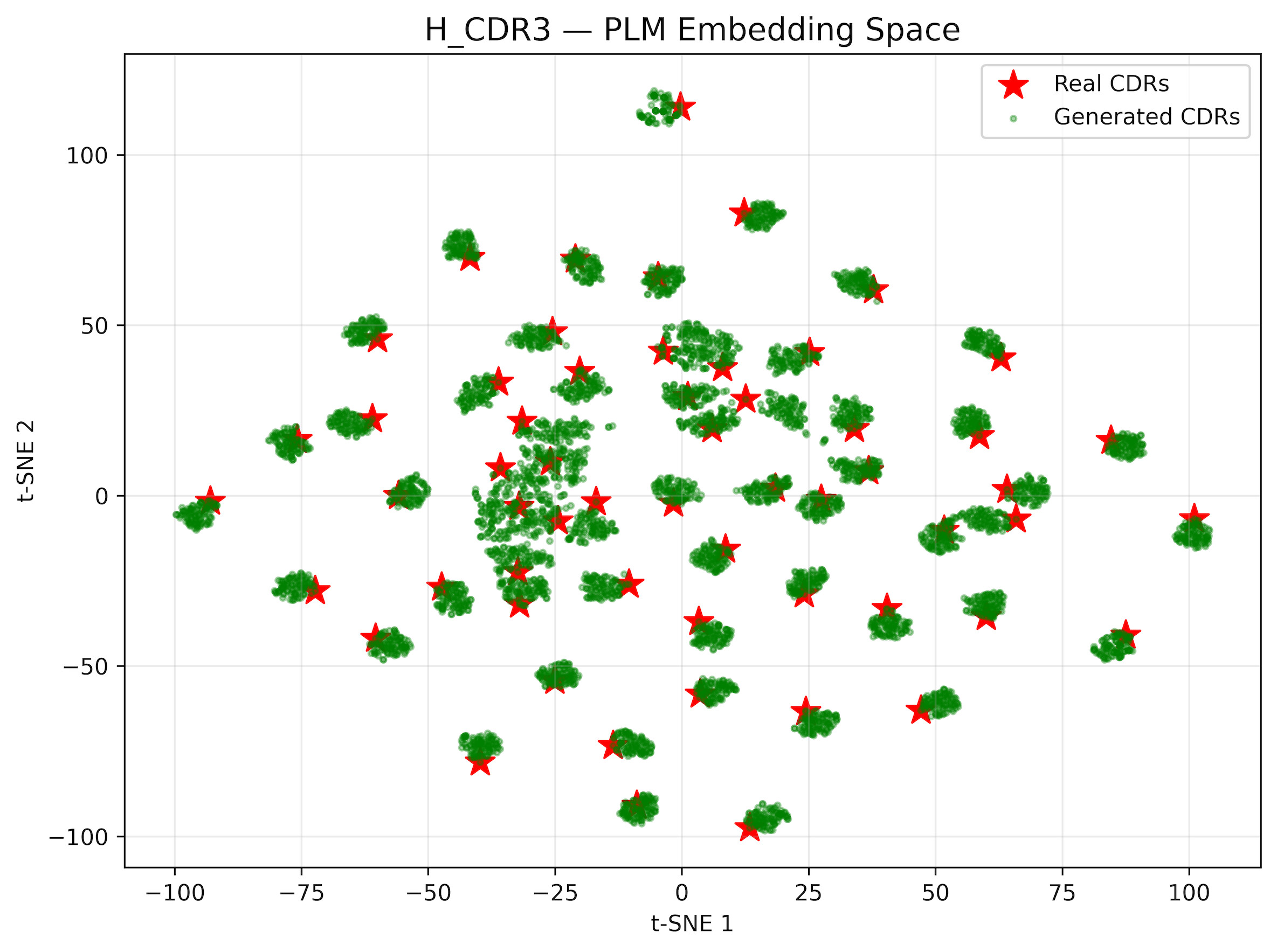}
        \caption{CDR-H3}
    \end{subfigure}

    \vskip\baselineskip

    \begin{subfigure}[b]{0.3\textwidth}
        \centering
        \includegraphics[width=\textwidth]{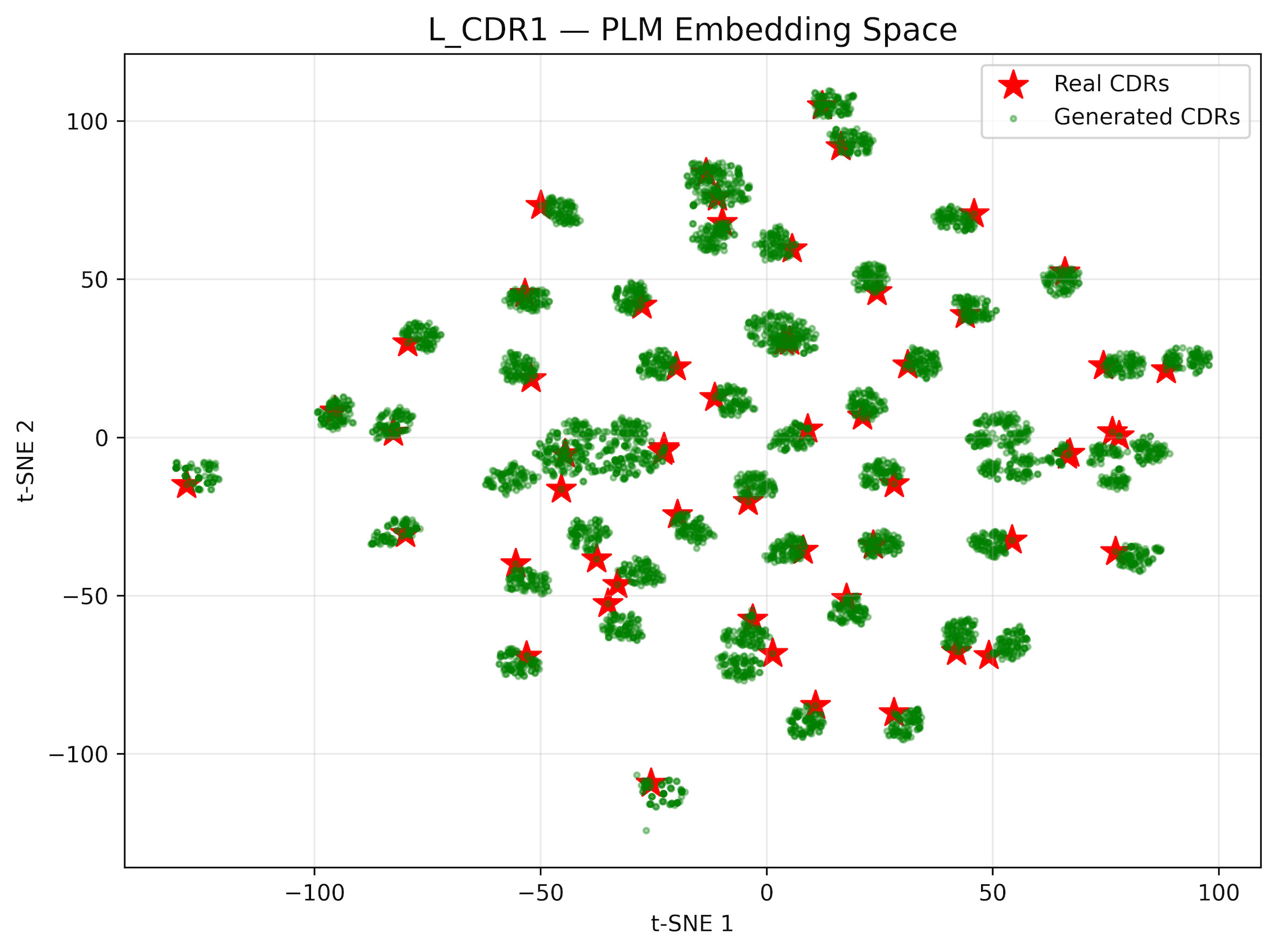}
        \caption{CDR-L1}
    \end{subfigure}
    \hfill
    \begin{subfigure}[b]{0.3\textwidth}
        \centering
        \includegraphics[width=\textwidth]{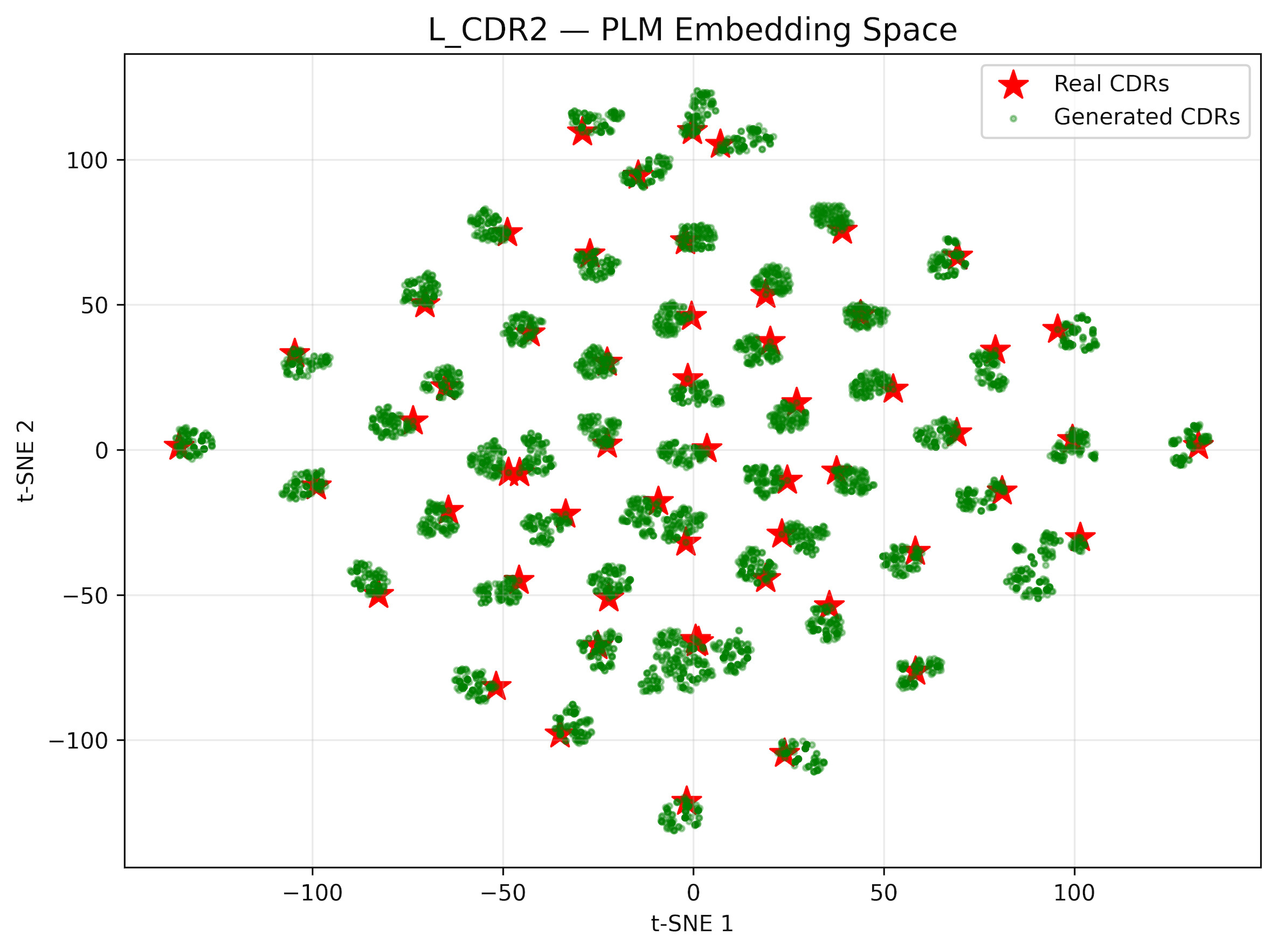}
        \caption{CDR-L2}
    \end{subfigure}
    \hfill
    \begin{subfigure}[b]{0.3\textwidth}
        \centering
        \includegraphics[width=\textwidth]{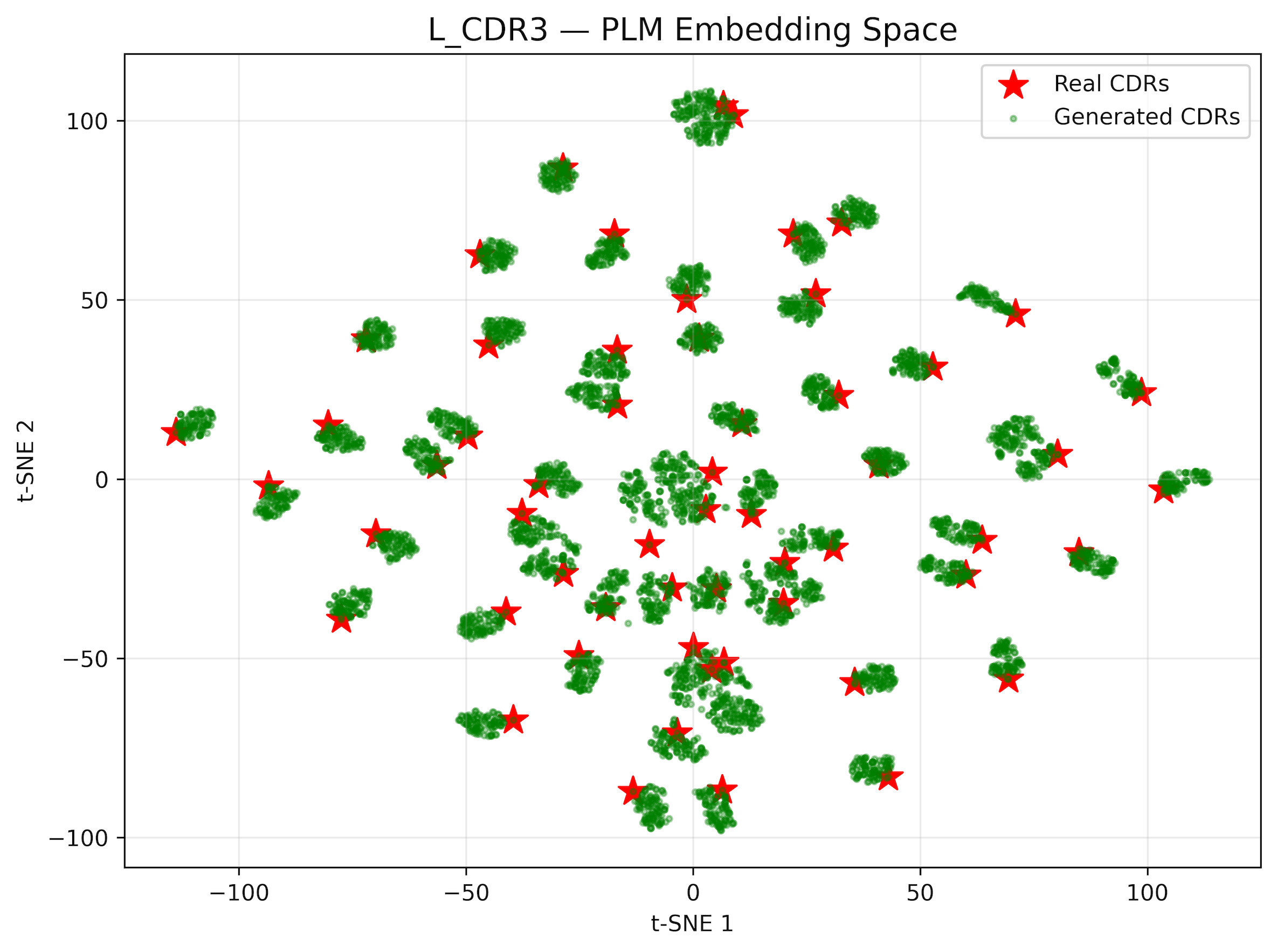}
        \caption{CDR-L3}
    \end{subfigure}

    \caption{tSNE clustering of ESM-2 embeddings of CDRs generated by AbFlowNet and the reference CDR of the RAbD dataset. The reference CDR is generally in the same cluster as the generated CDRs, showing that these are evolutionarily plausible and possess natural-like sequence properties according to ESM-2.}
    \label{fig:plm-embeddings}
\end{figure}

Protein Language Models (PLM) have been shown to learn evolutionary plausibility of protein sequences \citep{esm2} and have been successfully used in wet-lab applications \citep{plmuse}. We use the ESM-2 PLM to extract embeddings of the generated CDRs and compare them to the embeddings of the reference CDR to determine their evolutionary plausibility. Since the CDR regions are small, we use Locality-Aware Pooling \citep{hoang2025locality} to extract the CDR's embedding conditioned on the local antibody chain. Figure \ref{fig:plm-embeddings} shows that generated CDR sequences strongly cluster together with their corresponding reference sequences, showing that the generated CDRs are evolutionarily plausible and highly similar to their references according to ESM-2.

\section{Distribution of CDR Energy Metrics}
\label{app:distribution}
\begin{figure}[!htbp]
    \centering
    \caption{Distribution of sample CDRs generated by AbFlowNet on the RAbD test dataset.}
    \begin{subfigure}{\textwidth}
    \centering
    \caption{\texttt{\textbf{3bn9 complex}}}
    \includegraphics[width=0.95\textwidth]{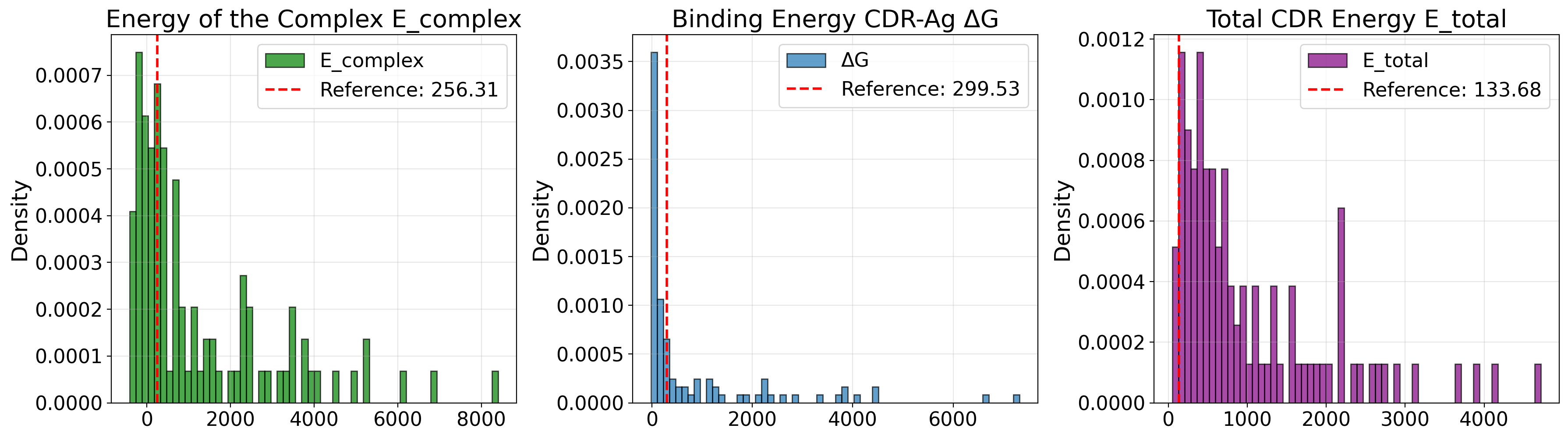}
    \end{subfigure}

    \begin{subfigure}{\textwidth}
    \centering
    \caption{\texttt{\textbf{2b2x complex}}}
    \includegraphics[width=0.95\textwidth]  {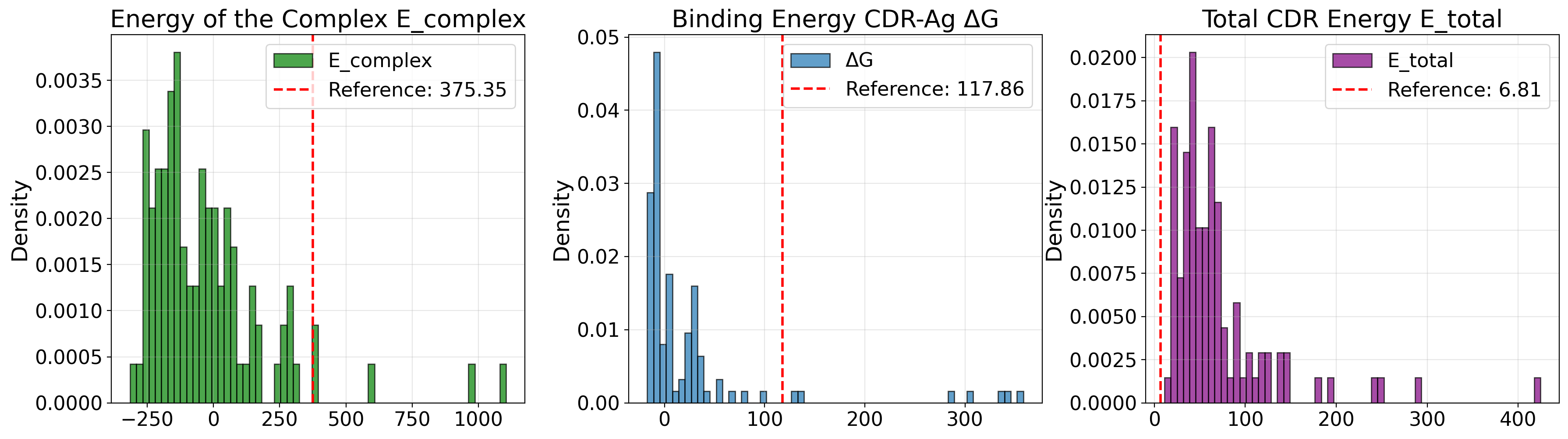}
    \end{subfigure}

    \begin{subfigure}{\textwidth}
    \centering
    \caption{\texttt{\textbf{4lvn complex}}}
    \includegraphics[width=0.95\textwidth]{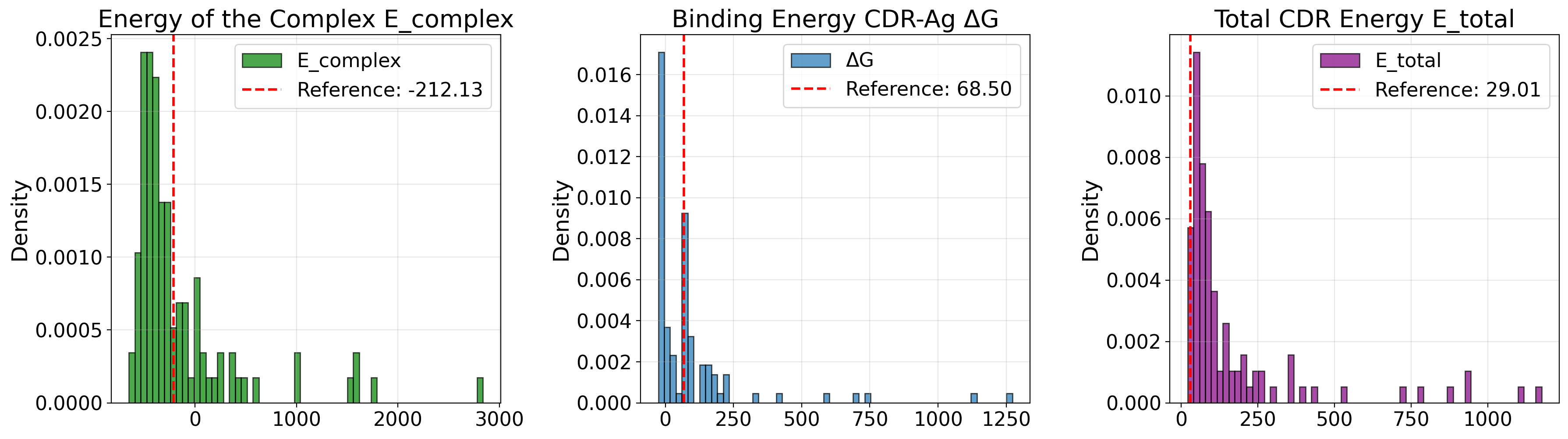}
    \end{subfigure}

    \begin{subfigure}{\textwidth}
    \centering
    \caption{\texttt{\textbf{2ghw complex}}}
    \includegraphics[width=0.95\textwidth]{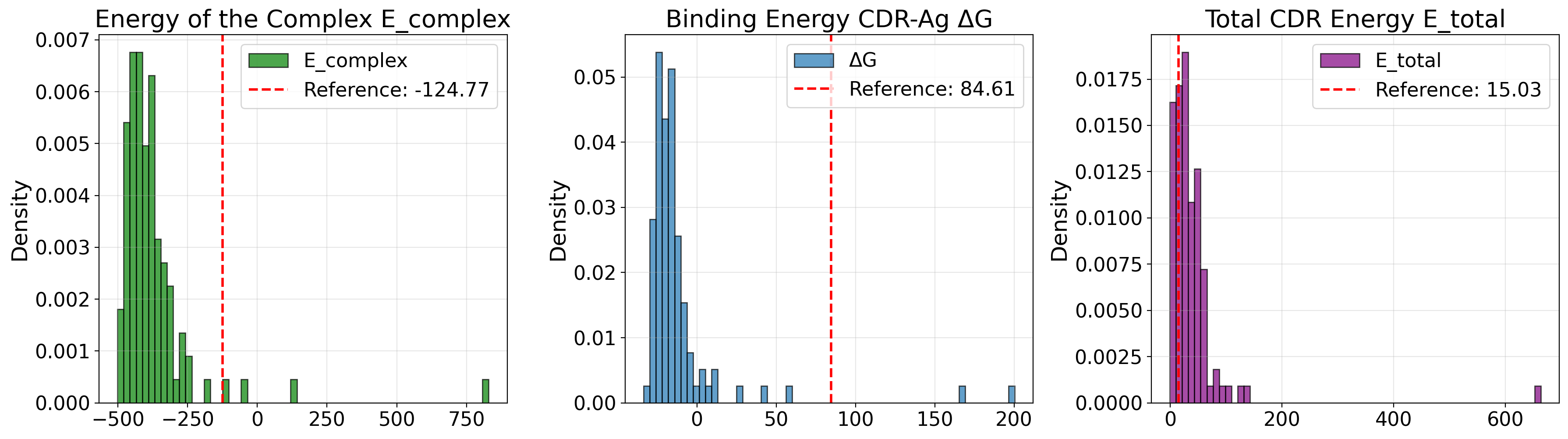}
    \end{subfigure}
\end{figure}

\begin{table}[htbp]
\centering
\caption{Mean, standard deviation, and minimum of Energy Metrics of CDRs generated by AbFlowNet on the RAbD test dataset. $\dagger$ denotes complexes that could not be processed with PyRosetta FastRelax.}
\resizebox{1\textwidth}{!}{
\begin{tabular}{l ccc ccc ccc}
\toprule
\multirow{2}{*}{\textbf{Complex}} & \multicolumn{3}{c}{\textbf{Energy of Complex $E_{\text{complex}}$ (↓)}} & \multicolumn{3}{c}{\textbf{Binding Energy CDR-Ag $\Delta G$ (↓)}} & \multicolumn{3}{c}{\textbf{Total CDR Energy $E_{\text{total}}$ (↓)}} \\
\cmidrule(lr){2-4} \cmidrule(lr){5-7} \cmidrule(lr){8-10}
 & \textbf{Min} & \textbf{Mean $\pm$ Std} & \textbf{Ref} & \textbf{Min} & \textbf{Mean $\pm$ Std} & \textbf{Ref} & \textbf{Min} & \textbf{Mean $\pm$ Std} & \textbf{Ref} \\
\midrule
1a14\_H\_L\_N & 465.24 & 739.27 $\pm$ 349.05 & 1829.81 & -23.46 & 5.76 $\pm$ 154.80 & 185.48 & 18.46 & 151.25 $\pm$ 168.06 & 56.52 \\
1a2y\_B\_A\_C & -695.23 & -643.06 $\pm$ 86.15 & -412.00 & -20.92 & -15.31 $\pm$ 2.99 & 17.32 & -9.88 & 10.63 $\pm$ 12.45 & -13.09 \\
1fe8\_H\_L\_A $\dagger$& 66.58 & 2244.82 $\pm$ 1279.23 & -300.06 & 151.96 & 1998.45 $\pm$ 1199.38 & 62.53 & 212.53 & 1444.82 $\pm$ 731.12 & 29.13 \\
1ic7\_H\_L\_Y $\dagger$ & 94.59 & 920.13 $\pm$ 702.52 & 29.49 & 54.26 & 250.78 $\pm$ 308.75 & 75.85 & 54.36 & 541.44 $\pm$ 397.55 & 17.72 \\
1iqd\_B\_A\_C & -591.29 & -464.97 $\pm$ 75.50 & -60.88 & -37.13 & 16.03 $\pm$ 35.73 & 206.79 & -1.97 & 33.88 $\pm$ 21.23 & 41.34 \\
1n8z\_B\_A\_C & 64.03 & 258.41 $\pm$ 166.25 & 2867.82 & 10.21 & 82.56 $\pm$ 82.50 & 188.81 & 8.13 & 95.22 $\pm$ 72.46 & 34.61 \\
1ncb\_H\_L\_N & 567.36 & 1235.35 $\pm$ 781.95 & 1602.17 & -4.30 & 128.75 $\pm$ 273.55 & 188.93 & 27.20 & 241.38 $\pm$ 337.78 & 31.78 \\
1osp\_H\_L\_O & -396.07 & -69.62 $\pm$ 279.88 & -10.07 & -18.15 & 61.66 $\pm$ 183.67 & 75.33 & 20.21 & 149.73 $\pm$ 135.03 & -2.64 \\
1uj3\_B\_A\_C & -611.02 & -250.98 $\pm$ 309.79 & -214.20 & -24.65 & 26.25 $\pm$ 58.20 & 54.90 & -1.91 & 36.66 $\pm$ 38.08 & -4.91 \\
2adf\_H\_L\_A & -519.56 & -283.61 $\pm$ 196.12 & -591.05 & -16.85 & 54.62 $\pm$ 127.52 & 25.70 & 10.47 & 66.50 $\pm$ 34.71 & -12.68 \\
2b2x\_H\_L\_A & -314.15 & -26.41 $\pm$ 231.04 & 375.35 & -17.12 & 24.69 $\pm$ 75.21 & 117.86 & 11.17 & 70.64 $\pm$ 60.82 & 6.81 \\
2cmr\_H\_L\_A & -581.82 & -355.59 $\pm$ 240.43 & 131.70 & -18.15 & 90.94 $\pm$ 169.18 & 106.36 & 6.15 & 80.40 $\pm$ 77.97 & -3.14 \\
2dd8\_H\_L\_S & -17.64 & 230.40 $\pm$ 155.19 & 424.31 & -93.32 & -8.70 $\pm$ 20.69 & 181.95 & -1.80 & 38.35 $\pm$ 36.71 & 74.37 \\
2ghw\_B\_b\_A & -501.10 & -373.60 $\pm$ 151.76 & -124.77 & -33.58 & -11.66 $\pm$ 31.11 & 84.61 & -0.74 & 39.36 $\pm$ 67.84 & 15.03 \\
2vxt\_H\_L\_I & -371.27 & 686.03 $\pm$ 1004.31 & -426.22 & 26.07 & 599.32 $\pm$ 838.56 & 32.43 & 25.65 & 641.44 $\pm$ 579.75 & -5.41 \\
2xqy\_G\_L\_A $\dagger$ & -805.28 & -673.89 $\pm$ 217.51 & -61.00 & -20.04 & -2.50 $\pm$ 8.44 & 33.51 & -5.38 & 36.07 $\pm$ 32.74 & -2.43 \\
2xwt\_A\_B\_C & -793.86 & -472.31 $\pm$ 246.79 & -614.54 & 16.28 & 132.46 $\pm$ 60.82 & 21.49 & 2.00 & 66.14 $\pm$ 51.55 & -14.67 \\
2ypv\_H\_L\_A & -854.85 & -705.91 $\pm$ 116.26 & -589.38 & -18.71 & 10.92 $\pm$ 24.19 & 37.34 & 7.40 & 52.64 $\pm$ 39.39 & 5.15 \\
3bn9\_D\_C\_B & -406.84 & 1233.60 $\pm$ 1772.78 & 256.31 & -16.47 & 815.04 $\pm$ 1437.08 & 299.53 & 52.33 & 1029.08 $\pm$ 985.85 & 133.68 \\
3cx5\_J\_K\_E & -175.80 & 154.47 $\pm$ 320.10 & 265.42 & -14.12 & 5.47 $\pm$ 25.08 & 60.79 & 21.83 & 171.98 $\pm$ 154.79 & -10.61 \\
3h3b\_c\_C\_B & -458.08 & -233.14 $\pm$ 277.84 & -11.97 & 2.34 & 6.65 $\pm$ 4.86 & 41.78 & 11.24 & 96.08 $\pm$ 84.23 & 10.90 \\
3hi6\_X\_Y\_B & -580.80 & -180.53 $\pm$ 502.57 & -374.75 & -17.26 & 138.10 $\pm$ 403.76 & 49.91 & 20.04 & 181.51 $\pm$ 198.83 & 5.80 \\
3k2u\_H\_L\_A & 2288.98 & 2402.07 $\pm$ 83.48 & 2858.45 & -26.46 & 3.46 $\pm$ 19.54 & 110.99 & 9.60 & 79.59 $\pm$ 58.88 & 18.79 \\
3l95\_B\_A\_X & -92.20 & 164.80 $\pm$ 119.44 & 219.03 & -12.62 & 27.84 $\pm$ 49.06 & 149.92 & 12.96 & 85.10 $\pm$ 45.78 & 2.93 \\
3mxw\_H\_L\_A & -483.43 & -339.37 $\pm$ 120.16 & -485.02 & -17.19 & -0.88 $\pm$ 16.07 & 26.31 & 6.20 & 55.40 $\pm$ 36.81 & -1.16 \\
3nid\_H\_L\_AD & -1627.05 & -1420.20 $\pm$ 157.44 & -1232.12 & -22.66 & 30.28 $\pm$ 25.74 & -1.90 & 16.14 & 119.02 $\pm$ 81.13 & -14.81 \\
3o2d\_H\_L\_A & -647.48 & -320.46 $\pm$ 339.92 & 99.11 & -108.43 & 19.95 $\pm$ 87.34 & 116.32 & 15.60 & 156.66 $\pm$ 162.21 & 0.16 \\
3rkd\_H\_L\_A & -559.58 & 546.00 $\pm$ 1289.63 & -293.93 & -26.78 & 237.52 $\pm$ 414.37 & -18.92 & 27.52 & 541.60 $\pm$ 615.68 & -2.53 \\
3s35\_H\_L\_X & -645.40 & -563.05 $\pm$ 80.61 & -123.47 & -20.89 & 14.90 $\pm$ 62.73 & 65.90 & -1.80 & 31.35 $\pm$ 29.16 & 1.75 \\
3uzq\_A\_a\_B & -544.58 & 442.73 $\pm$ 2213.79 & 27501.00 & -42.78 & -2.51 $\pm$ 123.08 & - & 7.26 & 191.34 $\pm$ 260.35 & -15.86 \\
3w9e\_A\_B\_C & -545.74 & -72.71 $\pm$ 407.60 & -129.53 & 2.03 & 100.45 $\pm$ 240.09 & 88.40 & 36.48 & 210.11 $\pm$ 162.17 & -2.35 \\
4cmh\_B\_C\_A & -836.91 & -441.40 $\pm$ 309.48 & -591.38 & -30.57 & 16.98 $\pm$ 47.83 & 51.93 & 18.21 & 187.31 $\pm$ 134.80 & -10.43 \\
4dtg\_H\_L\_K & -300.15 & -99.86 $\pm$ 192.58 & -125.16 & -6.02 & 18.94 $\pm$ 25.45 & 58.17 & 9.78 & 127.47 $\pm$ 144.68 & 15.80 \\
4dvr\_H\_L\_G & -366.66 & -253.46 $\pm$ 123.38 & 146.55 & -19.77 & -2.01 $\pm$ 38.03 & 58.58 & -8.60 & 40.92 $\pm$ 53.03 & -1.64 \\
4etq\_H\_L\_C & -564.55 & -337.49 $\pm$ 177.81 & -161.47 & -0.33 & 115.77 $\pm$ 50.70 & 73.30 & -3.46 & 41.51 $\pm$ 38.69 & -2.37 \\
4ffv\_D\_C\_B & -610.13 & -399.67 $\pm$ 131.91 & 347.73 & -20.34 & -6.64 $\pm$ 13.72 & 90.90 & 6.48 & 31.31 $\pm$ 28.74 & 26.73 \\
4fqj\_H\_L\_A & -191.74 & 1270.81 $\pm$ 1615.52 & 329.45 & 2.09 & 662.42 $\pm$ 998.74 & 130.07 & 75.53 & 732.24 $\pm$ 771.87 & 21.99 \\
4g6j\_H\_L\_A & -284.25 & -177.17 $\pm$ 106.34 & -278.38 & 2.98 & 39.60 $\pm$ 32.32 & 97.63 & -4.55 & 43.90 $\pm$ 31.31 & 7.59 \\
4g6m\_H\_L\_A & -727.04 & -621.49 $\pm$ 122.25 & -416.44 & -16.21 & 3.93 $\pm$ 29.46 & 36.74 & -7.36 & 42.04 $\pm$ 41.27 & -7.57 \\
4h8w\_H\_L\_G & -765.55 & -525.18 $\pm$ 167.65 & -500.04 & -22.41 & 62.50 $\pm$ 97.84 & 49.32 & 11.28 & 104.24 $\pm$ 83.04 & 1.33 \\
4ki5\_E\_F\_M & -193.51 & 289.45 $\pm$ 524.73 & -175.88 & -22.73 & 129.03 $\pm$ 232.47 & 45.83 & 22.34 & 211.71 $\pm$ 275.09 & -6.51 \\
4lvn\_C\_B\_A & -650.42 & -148.01 $\pm$ 571.38 & -212.13 & -25.83 & 94.06 $\pm$ 208.54 & 68.50 & 20.18 & 170.38 $\pm$ 233.80 & 29.01 \\
4ot1\_H\_L\_A & -377.87 & 1924.70 $\pm$ 2775.83 & -332.76 & -5.03 & 489.36 $\pm$ 1109.65 & 62.94 & 95.39 & 1792.68 $\pm$ 2029.26 & -2.69 \\
4qci\_B\_A\_D & -299.23 & -130.16 $\pm$ 123.36 & -76.90 & -17.54 & 53.22 $\pm$ 52.44 & 37.63 & 5.22 & 63.86 $\pm$ 63.54 & 14.68 \\
4xnq\_B\_A\_D & 190.39 & 582.77 $\pm$ 311.38 & 238.64 & -10.39 & 54.82 $\pm$ 71.60 & 34.20 & 41.31 & 271.33 $\pm$ 197.50 & -9.83 \\
4ydk\_H\_L\_G & -754.23 & 1338.02 $\pm$ 2353.61 & -617.74 & -25.80 & 841.97 $\pm$ 1496.09 & 22.24 & 73.71 & 1361.75 $\pm$ 1483.04 & -23.96 \\
5b8c\_B\_A\_C & -405.83 & -89.10 $\pm$ 259.21 & -322.93 & 15.95 & 119.52 $\pm$ 80.35 & 90.55 & 20.50 & 120.70 $\pm$ 107.06 & 0.10 \\
5bv7\_C\_B\_A & 196.59 & 920.92 $\pm$ 828.95 & 1230.31 & 21.23 & 172.09 $\pm$ 217.10 & 113.39 & 49.49 & 447.16 $\pm$ 464.02 & 44.74 \\
5d93\_C\_B\_A $\dagger$& -479.00 & 805.38 $\pm$ 1122.77 & -541.29 & 25.53 & 729.50 $\pm$ 854.56 & 22.08 & 29.29 & 860.37 $\pm$ 716.65 & -0.84 \\
5d96\_J\_I\_D & -197.67 & -50.32 $\pm$ 191.34 & 295.45 & -15.31 & 23.77 $\pm$ 103.76 & 87.44 & 1.94 & 91.77 $\pm$ 100.97 & 11.20 \\
5en2\_A\_B\_C & -681.11 & -257.33 $\pm$ 683.81 & -532.16 & -27.48 & 44.84 $\pm$ 220.13 & 14.59 & 20.79 & 280.10 $\pm$ 378.01 & -12.95 \\
5f9o\_H\_L\_G & -748.61 & -416.83 $\pm$ 315.42 & 67.19 & -39.86 & 13.56 $\pm$ 64.36 & 88.09 & 12.70 & 168.53 $\pm$ 152.44 & 4.09 \\
5ggs\_A\_B\_Z & -412.43 & -123.40 $\pm$ 249.30 & -137.59 & -12.58 & 52.22 $\pm$ 82.59 & 56.20 & 11.03 & 129.07 $\pm$ 135.37 & -8.39 \\
5hi4\_H\_L\_A & -508.21 & -398.51 $\pm$ 114.40 & -215.09 & -20.36 & 8.50 $\pm$ 96.39 & -24.37 & 6.73 & 55.59 $\pm$ 33.81 & 8.09 \\
5j13\_C\_B\_A & -475.93 & -298.67 $\pm$ 151.07 & -365.81 & -18.03 & 1.64 $\pm$ 19.68 & 16.54 & 16.60 & 116.51 $\pm$ 103.25 & 5.91 \\
5l6y\_H\_L\_C & -487.69 & -190.69 $\pm$ 319.95 & -360.32 & -29.21 & 20.40 $\pm$ 68.66 & -3.57 & 21.19 & 164.42 $\pm$ 174.97 & 6.56 \\
5mes\_H\_L\_A & -594.11 & -422.81 $\pm$ 94.28 & -156.46 & -7.90 & -0.74 $\pm$ 7.36 & -7.93 & -2.23 & 43.10 $\pm$ 35.82 & -9.65 \\
5nuz\_A\_B\_C & -758.42 & -556.64 $\pm$ 239.14 & -674.95 & -31.56 & -1.03 $\pm$ 36.37 & -11.37 & 11.88 & 87.82 $\pm$ 70.17 & -9.43 \\
\bottomrule
\end{tabular}
}
\end{table}

\section{Additional Qualitative Examples and Interpretation}
\label{app:qual}

We visualize the generated CDRs using PyMol \citep{delano2002pymol}. The globular structure on the top is the molecular surface of the antigen, and the ribbon-like structure at the bottom is the antibody, with the CDR-H3 region colored in. The large arrows represent the start and end of the rest of the antibody sequence that we keep fixed. The antigen surface has grooves and cavities where the CDR region binds. Omitting salt-bridge and hydrogen bond interactions for simplicity, we color-code the hydrophilic and hydrophobic residues of the antigen and the CDR. Hydrophobic antigen residues are colored in orange, and hydrophobic CDR residues are colored in red. Likewise, yellow and green represent hydrophilic residues in the antigen and CDR, respectively. \textbf{Naturally, due to specificity, a stable configuration would have matching regions closer together, orange-red for hydrophobic and yellow-green for hydrophilic}. Another aspect of consideration is the Binding Surface Area (BSA), which measures how deeply inset the CDR region is in the antigen. Specificity, surface area, and other factors such as the existence of salt bridges and hydrogen bonds, together determine the overall stability and binding energy.

\begin{figure}[h]
  \centering
  \begin{subfigure}{0.24\textwidth}
    \includegraphics[width=\linewidth]{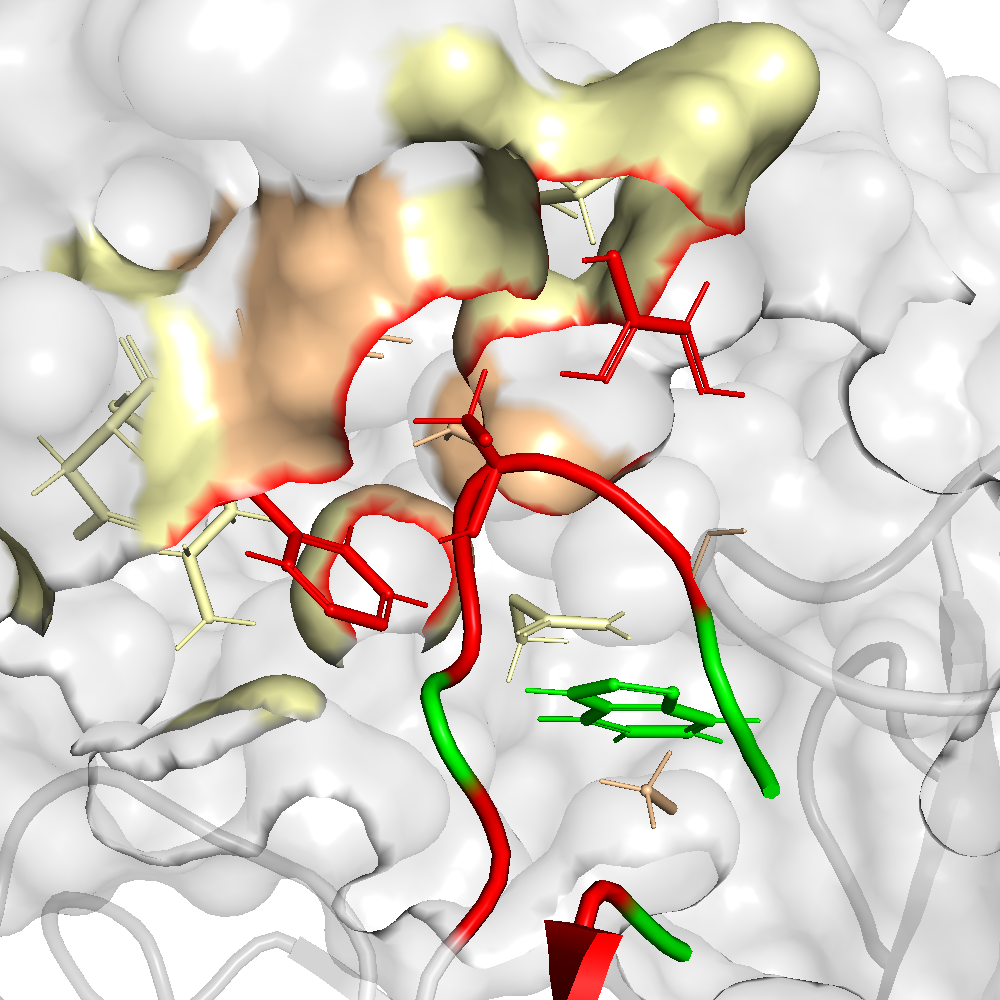}\captionsetup{justification=centering}
    \caption{DiffAb H3 \\$E_{total} =18.59 \text{ kcal/mol}$ \\$\Delta G=-2.52 \text{ kcal/mol}$}
  \end{subfigure}
  \hfill
  \begin{subfigure}{0.24\textwidth}
    \includegraphics[width=\linewidth]{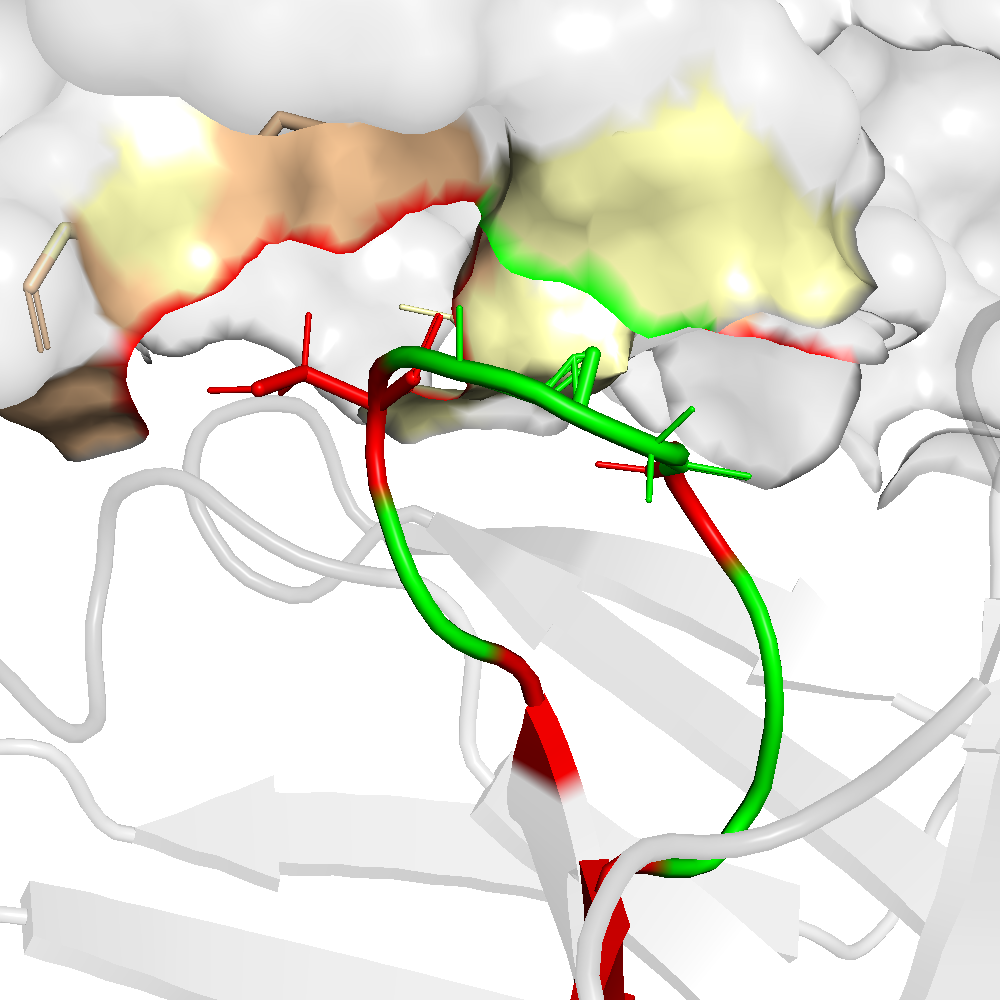}     \captionsetup{justification=centering}
    \caption{AbFlowNet H3 \\$E_{total} =-8.60 \text{ kcal/mol}$ \\$\Delta G=-19.79 \text{ kcal/mol}$}
  \end{subfigure}
  \hfill
  \begin{subfigure}{0.24\textwidth}
    \includegraphics[width=\linewidth]{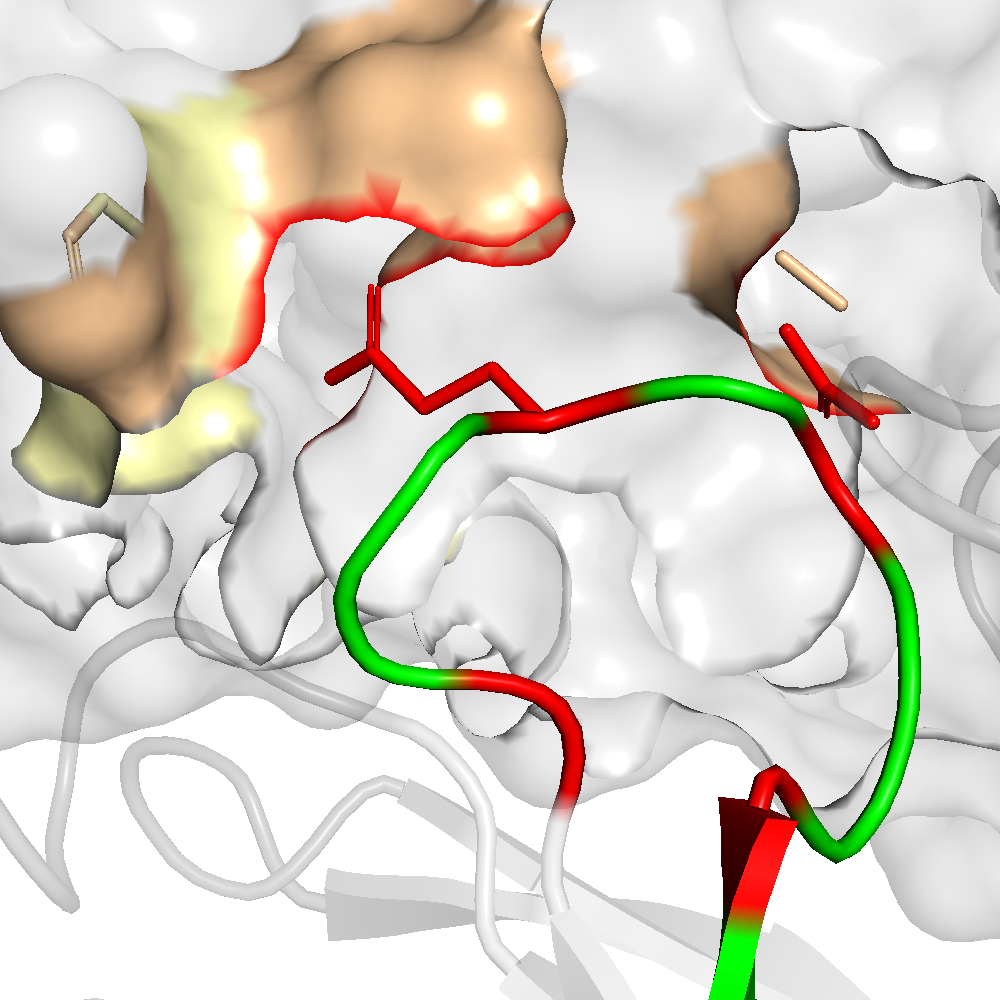}     \captionsetup{justification=centering}
    \caption{Reference H3 \\$E_{total} =-1.62 \text{ kcal/mol}$ \\$\Delta G=58.58 \text{ kcal/mol}$}
  \end{subfigure}
  \hfill
  \begin{subfigure}{0.24\textwidth}
   \includegraphics[width=\linewidth]{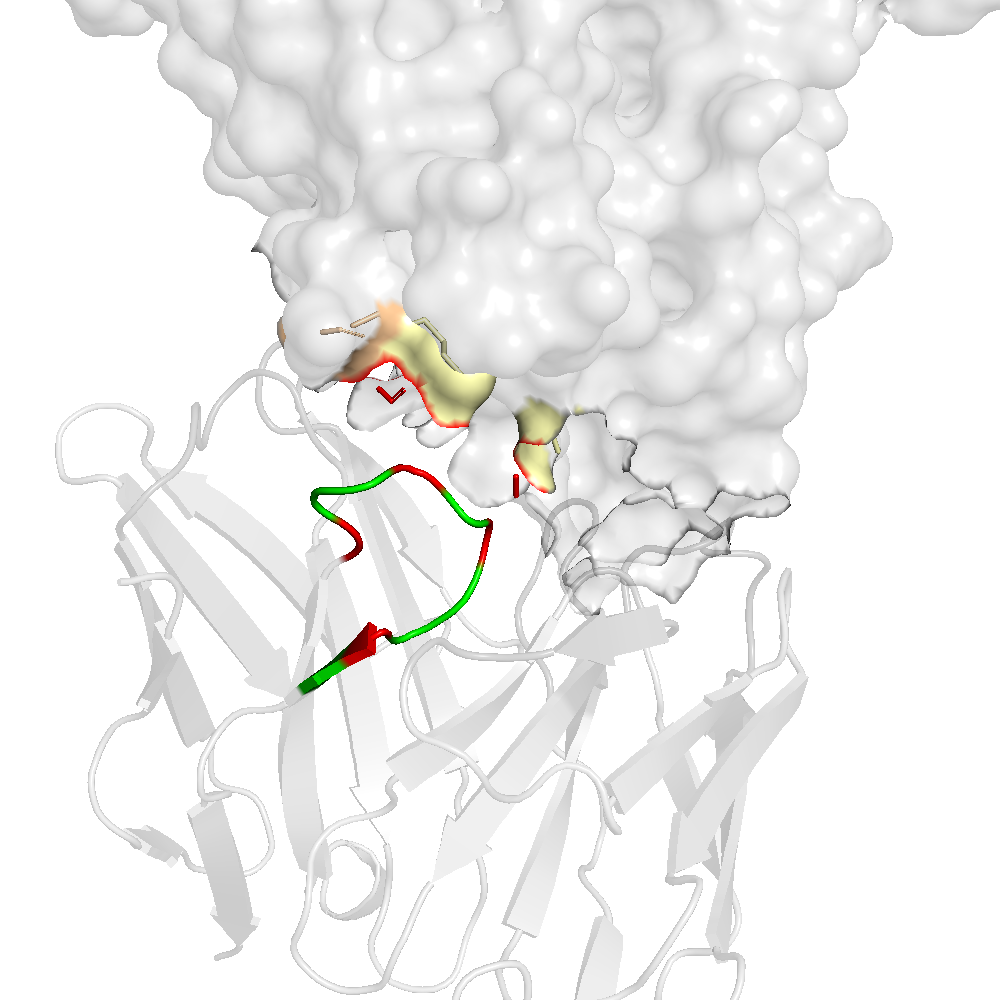}  \captionsetup{justification=centering}
    \caption{Reference \\ Antigen-Antibody\\Complex}
  \end{subfigure}
  \caption{\textit{De novo} Generated and Reference CDR-H3s for \texttt{4dvr} complex.  We selected the one with the highest $\Delta \text{G}$ out of 100 generations for DiffAb and AbFlownet.}
  \label{fig:4dvr}
\end{figure}

Figure \ref{fig:4dvr} highlights a case where AbFlowNet designs a CDR-H3 region that is embedded deeper in the antigen surface than the reference. The reference CDR-H3 region is further apart from the antigen surface and has a higher Binding Energy $\Delta \text{G}$ as a consequence. Both AbFlowNet and the reference have minor regions of poor specificity where hydrophobic and hydrophilic residues are close together (red-yellow interaction). In contrast, Diffab generates a CDR almost entirely with hydrophobic (red) residues and has large regions of poor specificity.

\begin{figure}[h]
  \centering
  \begin{subfigure}{0.24\textwidth}
    \includegraphics[width=\linewidth]{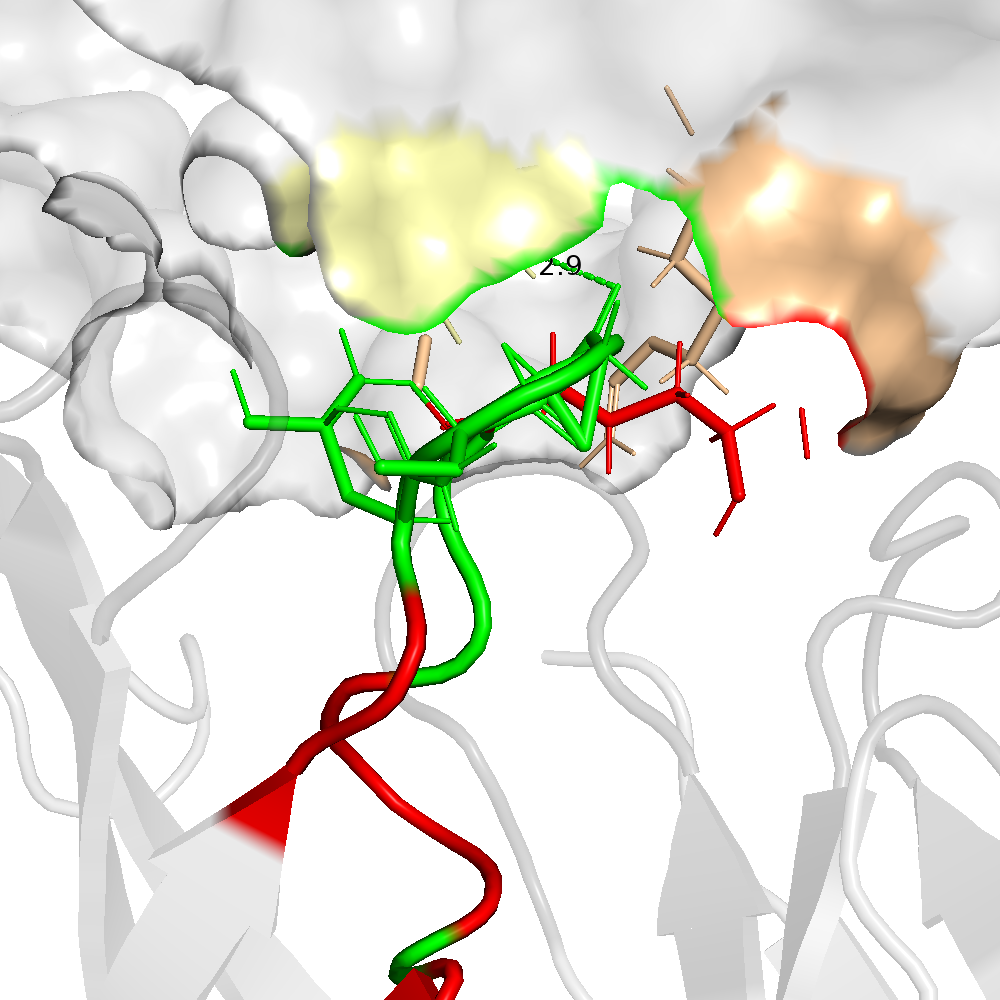}\captionsetup{justification=centering}
    \caption{DiffAb H3 \\$E_{total} = 49.60 \text{ kcal/mol}$ \\$\Delta G=4.44 \text{ kcal/mol}$}
  \end{subfigure}
  \hfill
  \begin{subfigure}{0.24\textwidth}
    \includegraphics[width=\linewidth]{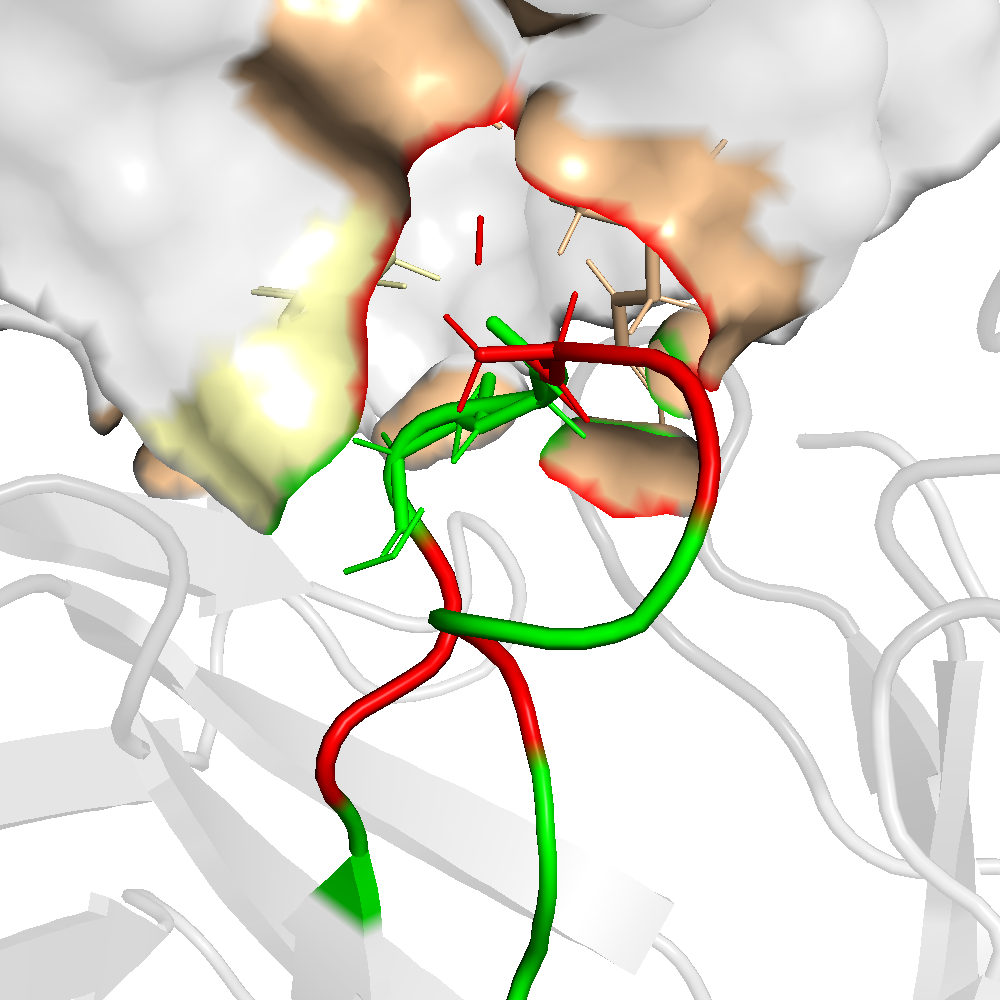}     \captionsetup{justification=centering}
    \caption{AbFlowNet H3 \\$E_{total} = 27.52 \text{ kcal/mol}$ \\$\Delta G=-26.78 \text{ kcal/mol}$}
  \end{subfigure}
  \hfill
  \begin{subfigure}{0.24\textwidth}
    \includegraphics[width=\linewidth]{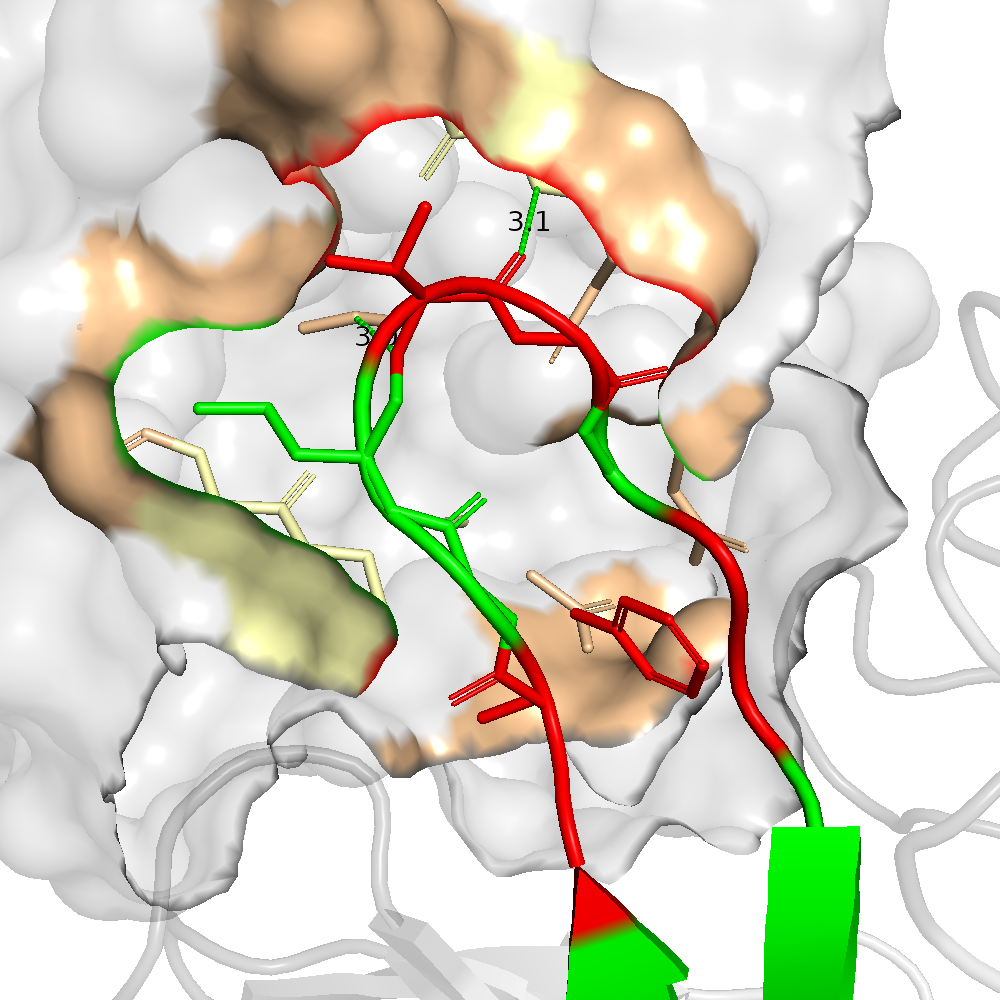}     \captionsetup{justification=centering}
    \caption{Reference H3 \\$E_{total} =-2.53 \text{ kcal/mol}$ \\$\Delta G=-18.92 \text{ kcal/mol}$}
  \end{subfigure}
  \hfill
  \begin{subfigure}{0.24\textwidth}
   \includegraphics[width=\linewidth]{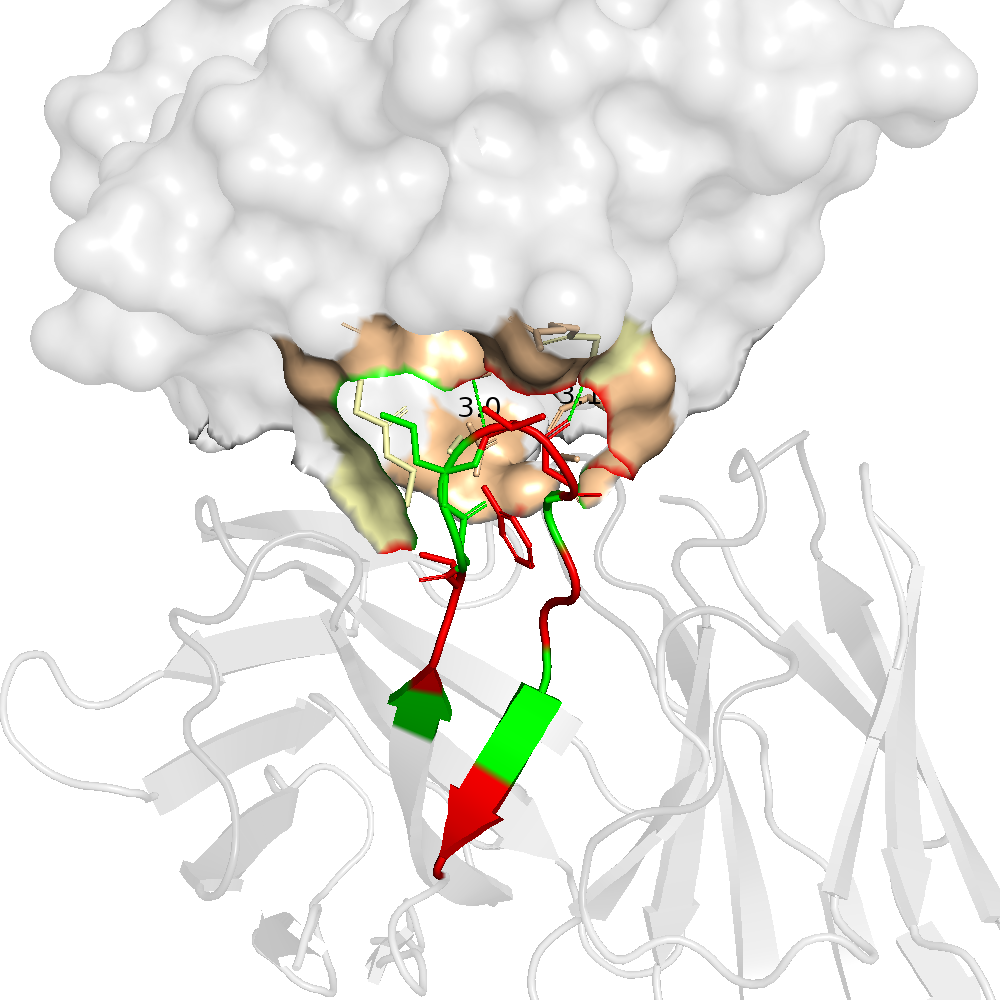}  \captionsetup{justification=centering}
    \caption{Reference \\ Antigen-Antibody\\Complex}
  \end{subfigure}
  \caption{\textit{De novo} Generated and Reference CDR-H3s for \texttt{3rkd} complex.}
  \label{fig:3rkd}
\end{figure}

Figure \ref{fig:3rkd} shows that the reference CDR-H3 for \texttt{3rkd} complex is embedded deeply in the antigen. Due to the deep inset, the reference is significantly lower, Total CDR Energy $E_{total} =-2.53$ than the generated CDRs. The CDR generated by AbFlowNet is also inset deeply but has a smaller binding surface area. Diffab's CDR-H3 is the furthest from the molecular surface of the antigen. The reference CDR has regions of poor specificity (orange-green interaction) while AbFlowNet shows high specificity.

\begin{figure}[h]
  \centering
  \begin{subfigure}{0.24\textwidth}
    \includegraphics[width=\linewidth]{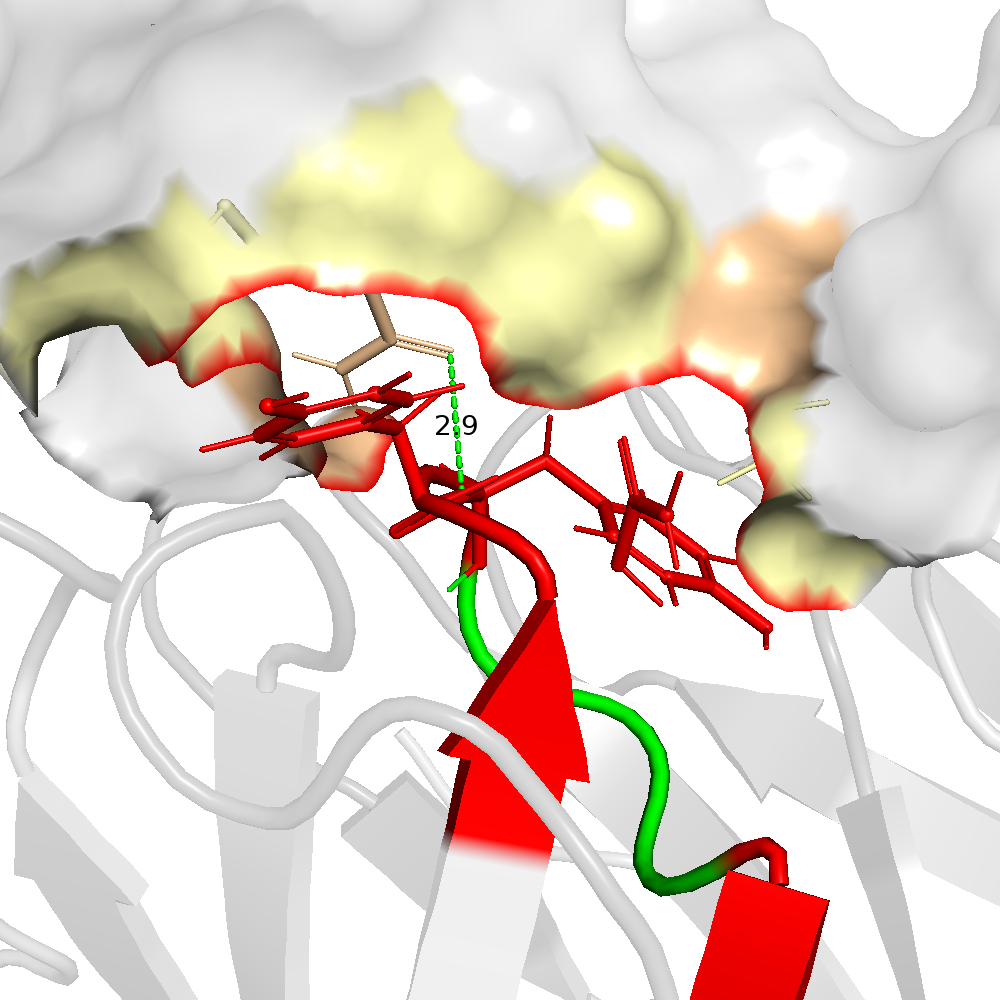}\captionsetup{justification=centering}
    \caption{DiffAb H3 \\$E_{total} = 16.62 \text{ kcal/mol}$ \\$\Delta G=-10.58 \text{ kcal/mol}$}
  \end{subfigure}
  \hfill
  \begin{subfigure}{0.24\textwidth}
    \includegraphics[width=\linewidth]{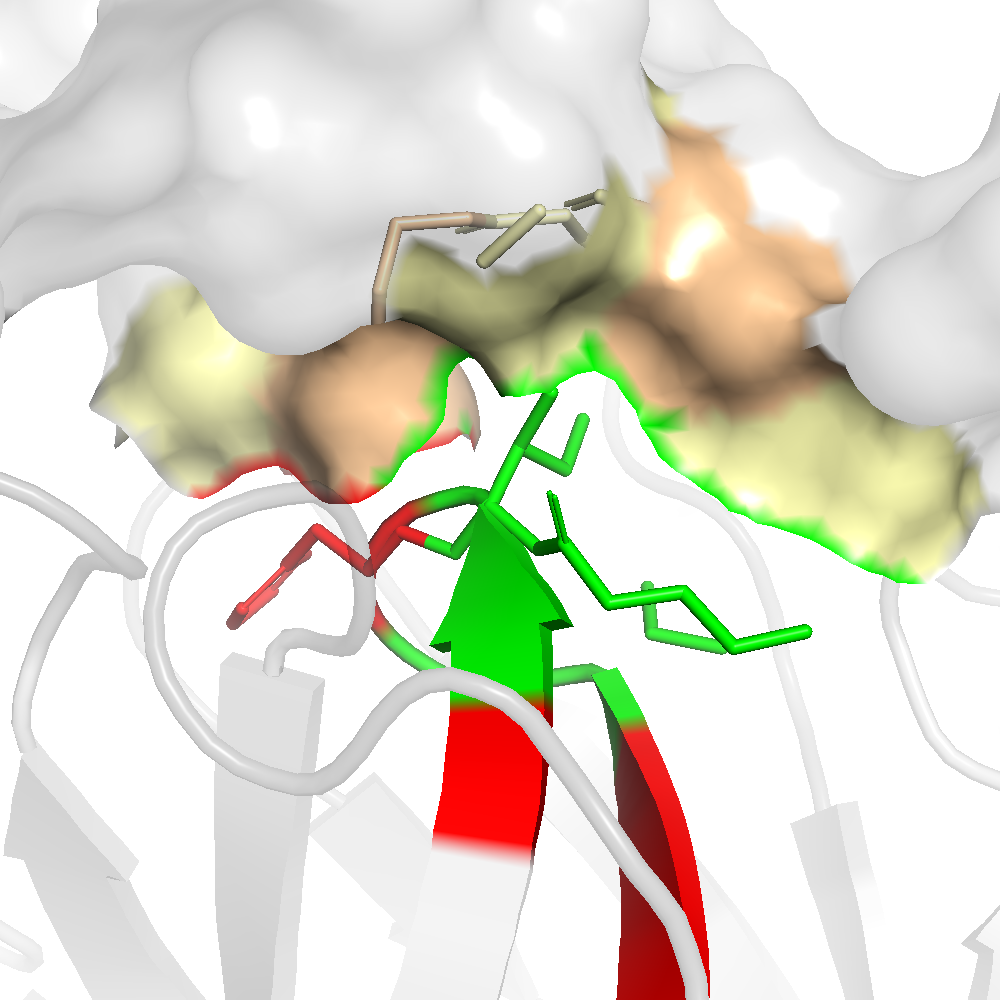}     \captionsetup{justification=centering}
    \caption{AbFlowNet H3 \\$E_{total} =6.73 \text{ kcal/mol}$ \\$\Delta G=-20.36 \text{ kcal/mol}$}
  \end{subfigure}
  \hfill
  \begin{subfigure}{0.24\textwidth}
    \includegraphics[width=\linewidth]{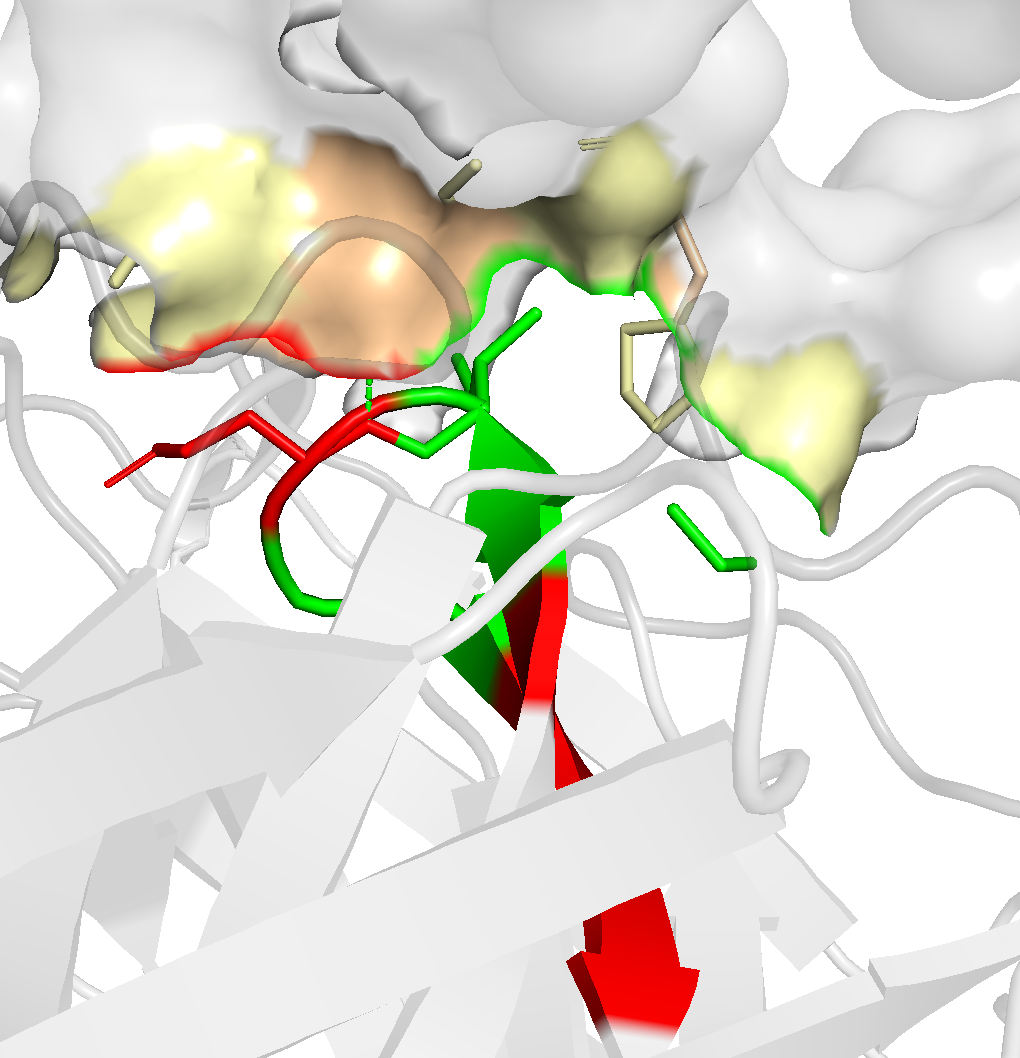}     \captionsetup{justification=centering}
    \caption{Reference H3 \\$E_{total} =8.09 \text{ kcal/mol}$ \\$\Delta G=-24.37 \text{ kcal/mol}$}
  \end{subfigure}
  \hfill
  \begin{subfigure}{0.24\textwidth}
   \includegraphics[width=\linewidth]{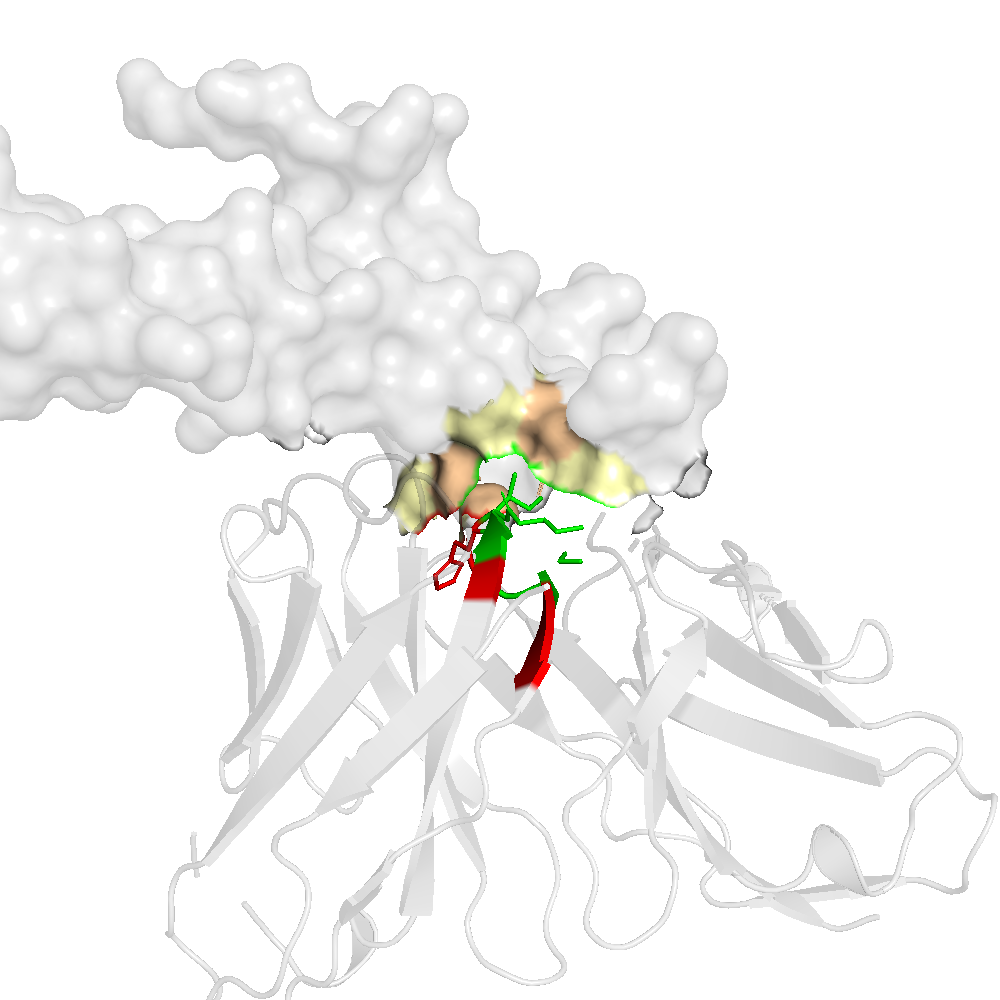}  \captionsetup{justification=centering}
    \caption{Reference \\ Antigen-Antibody\\Complex}
  \end{subfigure}
  \caption{\textit{De novo} Generated and Reference CDR-H3s for \texttt{5hi4} complex.}
  \label{fig:5hid}
\end{figure}

Figure \ref{fig:5hid} highlights a case where AbFlowNet generates a CDR-H3 very similar to the reference CDR-H3, differing only in one residue and minimal differences in position and orientation.  As such, $\Delta G$ and $E_{total}$ are very similar.  We note that our training set was deduplicated precisely based on $\geq 50\%$ similarity with the CDR-H3 region. This is a strong signal that AbFlowNet implicitly learns structures that optimize binding energy due to GFlowNet training. In contrast, similar to Figure \ref{fig:4dvr}, DiffAb once again generates a CDR-H3 region with only hydrophobic (red) residues and shows poor specificity (red hydrophobic CDR residues close to yellow hydrophilic antigen surface) despite having a deeper inset than even the reference.

\section{Comparison with Reward-Weighted Diffusion}
\label{app:reward_weighted}

In this section, we compare AbFlowNet with a simple reward-weighted baseline \citep{diffusionRL}, where the reconstruction loss $L = \mathbb{E}_{t \sim \text{Uniform}(1,\ldots,T)}\left[L^t_{\text{type}} + L^t_{\text{pos}} + L^t_{\text{ori}} \right]$ of each complex $S$ is weighted multiplicatively by $R(S) = \exp\!\left(-\alpha\,\Delta G(S)\right)$, normalized to unit mean within the mini-batch. Unlike AbFlowNet, where the TB loss objective is active for the last $5,000$ steps, reward-weighted baseline receives reward supervision for the entire duration of training. We additionally test the effect of additional training with the GFlowNet TB loss from the reward-weighted baseline, as denoted by RW + AbFlowNet in Table \ref{tab:reward-weighted}. We find that the reward-weighted baseline does not consistently improve over the DiffAb baseline, and both underperform AbFlowNet for all CDRs.

\begin{table}[ht]
\centering
\caption{Mean Top-1 CDR $\Delta G$ over RAbD test complexes per CDR. RW denotes Reward-weighted.}
\label{tab:reward-weighted}
\begin{tabular}{llr|llr}
\toprule
CDR & Method & $\Delta G$ & CDR & Method & $\Delta G$ \\
\midrule
\multirow{4}{*}{H-CDR1} & DiffAb & \underline{-58.18} & \multirow{4}{*}{L-CDR1} & DiffAb & \underline{-56.23} \\
 & RW & -57.97 &  & RW & -56.12 \\
 & AbFlowNet & \textbf{-58.98} &  & AbFlowNet & \textbf{-56.32} \\
 & RW + AbFlowNet & -57.40 &  & RW + AbFlowNet & -55.84 \\
\midrule
\multirow{4}{*}{H-CDR2} & DiffAb & \underline{-56.03} & \multirow{4}{*}{L-CDR2} & DiffAb & -56.29 \\
 & RW & -54.50 &  & RW & -56.35\\
 & AbFlowNet & \textbf{-58.23} &  & AbFlowNet & \underline{-57.00}\\
 & RW + AbFlowNet & -55.51 &  & RW + AbFlowNet & \textbf{-57.29} \\
\midrule
\multirow{4}{*}{H-CDR3} & DiffAb & -50.33 & \multirow{4}{*}{L-CDR3} & DiffAb & \underline{-55.87}\\
 & RW & -50.53 &  & RW & -54.50 \\
 & AbFlowNet & \underline{-53.39} &  & AbFlowNet & \textbf{-56.82}\\
 & RW + AbFlowNet & \textbf{-54.35} &  & RW + AbFlowNet & -55.19 \\
\bottomrule
\end{tabular}
\end{table}

\section{Acknowledgment Regarding Use of Large Language Models}

We used Large Language Models (LLMs) only as a writing aid to rephrase and correct the draft. We did not use LLMs to generate code, experiments, or ideation.

\section{Limitation}
\label{sec:limits}
1) The Trajectory Balance objective requires fully generating a CDR, which in our setup requires 100 forward passes with the neural network for each gradient update. 2) As the \textit{in vitro} affinity data for all training complexes are not available and for fair comparison with existing methods, we used \texttt{Pyrosetta InterfaceAnalyzer}, which is an unreliable estimator of binding energy. 3) Due to compute constraints, we sampled $100$ CDRs per complex in Table \ref{ssec:top_1}. Sampling at higher rates would potentially increase Top-1 metrics.

\end{document}